\documentclass[preprint,aps,pre,letterpaper,onecolumn,superscriptaddress]{revtex4-2} 

\usepackage{algorithm}
\usepackage{algpseudocode}
\usepackage[margin=1in]{geometry}
\usepackage{graphicx}
\usepackage{amssymb}
\usepackage{amsmath}
\usepackage{mathrsfs}
\usepackage{placeins}
\usepackage{caption3}
\usepackage{wrapfig}
\usepackage{caption} 
\usepackage{subcaption}
\usepackage{xcolor}
\usepackage{contour}
\usepackage{color}
\usepackage{outlines}
\usepackage[bookmarksnumbered=true,breaklinks=true,colorlinks,citecolor=blue,linkcolor=blue]{hyperref}

\usepackage[normalem]{ulem}
\definecolor{ao(english)}{rgb}{0.0, 0.5, 0.0}

\newcommand{\IM}{\mathcal{M}}

\newcommand{\reals}{\mathbb{R}} 

\newcommand{\trans}{^\mathrm{T}}
\newlength{\figwidth}
\newlength{\SCwidth}
\def\XXint#1#2#3{{\setbox0=\hbox{$#1{#2#3}{\int}$}
		\vcenter{\hbox{$#2#3$}}\kern-.5\wd0}}

\begin{document}


\title{Data-Driven Reduced-Order Modeling of Spatiotemporal Chaos with Neural Ordinary Differential Equations}


\author{Alec J. Linot}
\affiliation{Department of Chemical and Biological Engineering, University of Wisconsin-Madison, Madison WI 53706, USA}

\author{Michael D. Graham}
\email{mdgraham@wisc.edu}
\affiliation{Department of Chemical and Biological Engineering, University of Wisconsin-Madison, Madison WI 53706, USA}


\date{\today}

\begin{abstract}
Dissipative partial differential equations that exhibit chaotic dynamics tend to evolve to attractors that exist on finite-dimensional manifolds. We present a data-driven reduced order modeling method that capitalizes on this fact by finding the coordinates of this manifold and finding an ordinary differential equation (ODE) describing the dynamics in this coordinate system. The manifold coordinates are discovered using an undercomplete autoencoder -- a neural network (NN) that reduces then expands dimension. Then the ODE, in these coordinates, is approximated by a NN using the neural ODE framework. Both of these methods only require snapshots of data to learn a model, and the data can be widely and/or unevenly spaced. We apply this framework to the Kuramoto-Sivashinsky for different domain sizes that exhibit chaotic dynamics. With this system, we find that dimension reduction improves performance relative to predictions in the ambient space, where artifacts arise. Then, with the low-dimensional model, we vary the training data spacing and find excellent short- and long-time statistical recreation of the true dynamics for widely spaced data (spacing of $\sim$0.7 Lyapunov times). We end by comparing performance with various degrees of dimension reduction, and find a ``sweet spot" in terms of performance vs.\ dimension.
\end{abstract}


\maketitle


\section{Introduction} \label{sec:Intro}

A common question in many applications is: given a time series of measurements on a system how can a predictive model be generated to estimate future states of the system? For problems like weather forecasting, just knowing the future state of the system is useful. In other problems, like minimizing turbulent drag on an aircraft, models are desirable for creating control policies. These models can sometimes be generated from first principles, but insufficient information on the system often limits the ability to write them explicitly. Even when a model can be written out explicitly, it may be very high-dimensional and computationally expensive to simulate. Thus it is often desirable to generate a low-dimensional model from data.

We consider data sets $\left\{u(t_1), u(t_2), \ldots, u(t_N)\right\}$, where $u(t_i)\in \mathbb{R}^d$  comes from measuring the state of a system at a given time $t_i$. The full state at a given time can be represented by direct measurements (e.g.\ position and velocity of a pendulum) or by a representation that is diffeomorphic to the state. One such representation is time delay measurements of the system, as shown by Takens \citep{Takens}. In the case of the pendulum, for example, this could correspond to writing the state as $u(t)=[\theta(t),\theta(t-\tau),\theta(t-2\tau),\theta(t-3\tau)]$, where $\theta$ is the angle of the pendulum and $\tau$ is the delay time. If $u$ lies on a $d$-manifold then a time delay representation in $\mathbb{R}^{2d+1}$ is diffeomorphic to u \citep{Takens}.  

For many systems of interest the state is high-dimensional and comes from a chaotic time series \cite{Ivancevic2007}. 
Despite this complex behavior, when these systems are dissipative the long-time dynamics often lie on a smooth invariant finite-dimensional inertial manifold $\mathcal{M} \subset \mathbb{R}^d$ of a much lower dimension ($d_\mathcal{M}$) than that of the state \cite{Temam1990}. 
Two such systems are the Kuramoto-Sivashinsky equation (KSE) \cite{Foias1988a,Temam1994,Jolly2000,Zelik2014} and the complex Ginzburg-Landau equation \cite{Doering1988}. Similarly, for the Navier-Stokes in two and three spatial dimensions, it has been shown that there is an approximate inertial manifold \cite{Foias1988,Temam1989}. It is approximate in the sense that there are small variations in many dimensions that can be assumed to be zero and still provide an accurate approximation of the state.

With a mapping to the manifold coordinate system the state can be represented (at least locally) in the manifold coordinates $h(t)\in\mathbb{R}^{d_\IM}$. That is, mappings $\chi$ and $\check{\chi}$ exist such that  $h=\chi(u)$ and $u=\check{\chi}(h)$. 
In the machine learning literature, $\chi$ and $\check{\chi}$ correspond to the the encoder and decoder, respectively, of a so-called undercomplete autoencoder structure \cite{Linot2020}, as we further discuss below. It should be noted that  there is no guarantee that $\IM$ can be \emph{globally} represented with a cartesian representation in $d_\IM$ dimensions. Indeed in general this cannot be done, and an ``atlas" of overlapping local representations, or ``charts", must be used \cite{lee2003}.  The application considered here will not require this more general formalism, but for related work using charts, see \cite{floryan2021charts}. Our aim here is to use data from a spatiotemporally chaotic system to learn the mappings  $\chi(u)$ and $\check{\chi}(h)$ back and forth between $\reals^d$ and a coordinate system on $\IM$ and then to learn the evolution equation for the dynamics in this coordinate system. This will be a minimal-dimensional representation for the dynamics of the system in question. We first introduce some background for the dimension-reduction problem, and then for the time-evolution problem. 

Dimension-reduction is a challenging and widely-studied problem, and many approaches have been considered.  Frequently a linear dimension reduction technique is used (i.e.\ $\IM$ is taken to exist in a linear subspace of $\mathbb{R}^d$). This choice can be rationalized by invocation of Whitney's theorem \cite{Whitney1944}: any smooth $d_\IM$-manifold can be embedded into $\mathbb{R}^{2d_\IM}$.  Sauer et al.\ later refined this result by showing this embedding can be performed by almost every smooth map from a $d_\IM$-manifold to $\mathbb{R}^{n}$ for $n>2d_b$ \cite{Sauer1991}, where $d_b$ is the box-counting dimension of the attractor that lies on the manifold $\IM$. The box-counting dimension is a fractal dimension that must be less than the manifold dimension.
	Thus, for almost every smooth map, including linear ones, $h \in \mathbb{R}^{2d_\mathcal{M}+1}$ contains the same information as $u$ (i.e. a map exists for reconstructing $u$ from $h$). 

Cunningham and Ghahramani \cite{Cunningham2015} give an overview of linear dimension techniques from a machine learning perspective. Many of these collapse to one of the most common methods -- principal component analysis (PCA). In PCA the linear transformation that minimizes the reconstruction error or maximizes the variance in the data is found. This transformation comes from projecting data onto the leading left singular vectors $U\in\mathbb{R}^{d\times d_h}$ of the centered snapshot matrix $X=[u(t_0)-\left<u\right>,...,u(t_N)-\left<u\right>]$ (here $\left<.\right>$ denotes the mean). This is equivalent to finding the eigenvectors of the covariance matrix $X X^T$. Similarly, the eigenvectors of $X^T X$ can be found and related to the eigenvectors of $X X^T$ through the singular value decomposition. This approach is sometimes known as classical scaling \cite{VanDerMaaten2009}. The projection onto these modes is $\tilde{u}=U U^T u$, where $\sim$ denotes the approximation of the state $u$. In this projection, $h=U^T u$ is the low-dimensional representation. Although this projection is the optimal linear transformation for minimizing the reconstruction error, $h$ may still require as many as $2d_\mathcal{M}+1$ dimensions to contain the same information as $u$.

Finding a \emph{minimal} representation in general requires a nonlinear transformation. Many different techniques exist for nonlinear dimension reduction, often with the goal of preserving some property of the original dataset in the lower dimension. Some popular methods include kernel PCA, diffusion maps, local linear embedding (LLE), and t-distributed stochastic neighbor embedding (tSNE) \cite{VanDerMaaten2009}. In kernel PCA the matrix $X^T X$ is replaced with a kernel matrix $K(u_i,u_j)$. This can be viewed as an application of the ``kernel trick" on the the covariance matrix between data that has been mapped into an higher-dimensional (often infinite-dimensional) ``feature space" \cite{VanDerMaaten2009}. The low-dimensional representation is then computed using the eigenvectors such that $h=[\sum_i^d v_{i,1} K(u_i,u),...,\sum_i^d v_{i,d_h} K(u_i,u)]$, where $v_{i,k}$ is the \emph{i}th element of the \emph{k}th eigenvector of $K(u_i,u_j)$.

Similarly, diffusion maps and LLE can be viewed as extensions of kernel PCA with special kernels \cite{VanDerMaaten2009}. In diffusion maps a Gaussian kernel is applied to the data giving the matrix $A=\exp (-||u_i-u_j||/2\epsilon)$, where $\epsilon$ is a tuning parameter that represents the local connectivity \cite{Ferguson2011}. Then the dimension reduction is performed by computing the eigenvalue decomposition of this matrix (normalized so columns add to one) giving $h_i=[v_{i,2},...,v_{i,d_h+1}]$. The first eigenvector is the trivial all-ones vector \cite{Ferguson2011}.
In LLE, a linear combination of the $k$ nearest neighbors to a data point are used to minimize $\sum_i||u_i-\sum_jW_{i,j}u_j||^2$, where $W_{i,j}=0$ if $u_j$ is not a neighbor of $u_i$. Then the low-dimensional representation $h$ is calculated to minimize $\sum_i||h_i-\sum_jW_{i,j}h_j||^2$ using the $W$ calculated in the first step \cite{Roweis2323}. The solution that minimizes this cost function can be found by computing the eigenvectors of $(I-W)^T(I-W)$, and letting $h_i=[v_{i,1},...,v_{i,d_h}]$.

The last algorithm we mention here is tSNE \cite{Hinton2003}. Unlike the previous methods which use the eigenvalue decompositions to solve the optimization problem, tSNE uses gradient descent. In tSNE, the objective is to match the distribution of the data in the high-dimensional space to the data in the low-dimensional space. This is done by finding $h$ that minimizes the Kullback-Leibler (KL) divergence $\sum_i \sum_j p_{i,j} \log p_{i,j}/q_{i,j}$, where $p$ and $q$ are the distributions of the high- and low-dimensional data respectively. In tSNE these distributions are approximated by $p=\exp (-||u_i-u_j||^2/2\sigma^2)/\sum_{k\neq l}\exp (-||u_k-u_l||^2/2\sigma^2)$ and $q=(1+||h_i-h_j||^2)^{-1}/\sum_{k\neq l}(1+||h_k-h_l||^2)^{-1}$. 
There is no guarantee in finding a global minimum or in finding the same solution every time. tSNE is frequently used for visualizing high-dimensional data in two or three dimensions, because it often separates complex data sets out into visually distinct clusters.

A few major drawbacks of these nonlinear dimension reduction techniques are they do not provide the function $\check{\chi}$, which reconstructs $u$ from $h$, they do not provide a method for adding out-of-sample data, and they do not scale well for large amounts of data. Except for tSNE, the Nystr$\ddot{\text{o}}$m extension can be used to resolve the out-of-sample problem for these methods \cite{Bengio2004}. However, using the Nystr$\ddot{\text{o}}$m extension results in different functions for $\chi$ depending on whether the data is in-sample or out-of-sample.

Because of these limitations, instead of using one of the above techniques, we tackle the dimension reduction problem directly, by approximating the mappings to the manifold coordinates $\chi$ and back $\check{\chi}$ as NNs. I.e., we use an undercomplete autoencoder \cite{Hinton2006,IanGoodfellowYoshuaBengio2017}. 
We describe autoencoders in more detail in Section \ref{sec:Framework}.

We now turn to developing, from data, dynamical models for the evolution of a state $u(t)$.  We consider systems that display deterministic, Markovian dynamics, so if $u(t)$ is the full state (either directly or diffeomorphically through embedding of a sufficient number of time delays), then the
dynamics can either be represented by a discrete time map 
\begin{equation}\label{eq:discrete}
	u(t+\tau)=F(u(t)),
\end{equation}
or 
an ordinary differential equation (ODE)
\begin{equation}\label{eq:ODE}
	\dfrac{d u}{d t}=f(u).
\end{equation} 
The learning goal is an accurate representation of $F$ or $f$, or their lower-dimensional analogues.

We consider first the discrete-time problem. This is easier because that is the format in which we usually have data. For simple systems that decay to a fixed point or display quasiperiodic dynamics with discrete frequencies, linear discrete-time methods, like dynamic mode decomposition (DMD) \cite{DMDBook},
can predict dynamics. However, these models are unable to predict the long-time behavior of chaotic systems, which is our interest here.

Predicting chaotic dynamics requires a nonlinear model. Among the most successful methods for predicting chaotic dynamics are reservoir networks \cite{Pathak2018a} and recurrent neural networks (RNN) \cite{Vlachas2018,Vlachas2019}. These methods use discrete time maps, are non-Markovian, and typically increase the dimension of the state.
Reservoir networks work by evolving forward a high-dimensional reservoir state $r(t)\in \mathbb{R}^{d_r}$, and finding a mapping from $r$ to $u$. This reservoir state is evolved forward in time by some function $r(t+\tau)=G(r(t),W_\text{in}u(t))$, where $G$ and $W_\text{in}$ are chosen a priori. Then, the task is finding the optimal parameters $p$ in $\tilde{u}(t+\tau)=W_\text{out}(r(t+\tau);p)$ that minimize $\left<||u(t+\tau)-\tilde{u}(t+\tau)||^2\right>$. This is non-Markovian because the prediction of the state $u(t+\tau)$ depends on the previous $u(t)$ and $r(t)$.
 In Pathak et al.\ \cite{Pathak2018a} a reservoir network was used to predict the chaotic dynamics of the KSE. For data with a state dimension $d=64$, they choose a reservoir dimension $d_r=5000$.  That is, here  the dimension of the representation has not been reduced, but rather expanded, by two orders of magnitude.

Similar to reservoir networks, RNNs have a hidden state $h_r\in \mathbb{R}^{d_{h_r}}$ that is evolved forward in time. The hidden state is evolved forward by $h_r(t+\tau)=\sigma_h(u(t),h_r(t);W_h)$, and the future state is estimated from the hidden state $\tilde{u}(t+\tau)=\sigma_u(h_r(t+\tau);W_u)$. The functions $\sigma_h$ and $\sigma_u$ take different forms depending upon the type of RNN -- two examples are the long short-term memory (LSTM) \cite{Hochreiter1997} and the gated recurrent unit \cite{cho-etal-2014-properties}. Regardless of architecture, the functions $\sigma_h$ and $\sigma_u$ are constructed from NN with parameters $W_h$ and $W_u$. 
These parameters come from minimizing the same loss as in reservoir computing $\left<||u(t+\tau)-\tilde{u}(t+\tau)||^2\right>$. 
Vlachas et al.\ \cite{Vlachas2019} provide a comparison between reservoir networks and RNN for chaotic dynamics of the KSE. Both provide predictive capabilities for multiple Lyapunov times, but, as noted, are high-dimensional and non-Markovian. Additionally, these methods typically require evenly spaced data for training, and start up often requires multiple known states, instead of a single initial condition. 

In contrast to these high-dimensional non-Markovian approaches to learning a discrete time mapping we have shown in prior work \cite{Linot2020} that a dense NN can approximate $F(u)$. This approach is advantageous because, like the underlying system, it is Markovian -- predictions of the next state only depends on the current state. Unlike the previous methods, this approach also drastically reduced the dimension of the state by representing it in terms of the manifold coordinates. For example, at a domain size of $L=22$, for the KSE, we showed that the state $u\in\mathbb{R}^{64}$ could be represented in $h\in\mathbb{R}^8$. We found this coordinate system using an undercomplete autoencoder that corrected on PCA, as described in more detail in Section \ref{sec:autoencoders}. Representing the state in this minimal fashion allows for faster prediction, and may provide insight due to limiting the system to only the essential degrees of freedom.

Rather than approximating $F(u)$ in Eq.\ \ref{eq:discrete}, the discrete-time representation, one can, in principle, approximate $f(u)$ in Eq.\ \ref{eq:ODE} the continuous-time representation.
 This is more challenging, because one does not generally have access to data for $du/dt$. This can be estimated, of course, from data closely spaced in time.  
 When the time derivative is known (or estimated) for all of the states ($f(u_i)$), Gonzalez Garcia et al.\ \cite{Gonzalez-Garcia1998} showed that a dense NN can provide an accurate functional form of $f$ for the KSE in a regime that exhibits a periodic orbit. In their work they input the state and its derivatives into the NN.
When $f(u_i)$ is unknown, Raissi et al.\ \cite{Raissi2018} 
showed it can be approximated from data with a multistep time-integration scheme such as the second-order Adams-Bashforth formula, and a NN can be trained to match this approximation. 

To estimate $f$ from data that is not closely spaced, Chen et al.\ \cite{Chen2019} have introduced an approach, which they call ``neural ODEs", where they use a NN to compute $f$ at each step in an ODE solver. This allows them to compute the solution at arbitrary points in time. We further describe, and apply, this approach in Section \ref{sec:Framework}. To determine the parameters of $f$, the difference between data and predictions are minimized. To determine the derivatives of $f$ with respect to the NN parameters, either an adjoint problem can be solved, or an automatic differentiation method used. 

Neural ODEs have been applied to a number of time series problems.  
Maulik et al.\ \cite{MAULIK2020} used neural ODEs and LSTMs for Burgers equation, and showed both outperform a Galerkin projection approach. 
Portwood et al.\ \cite{portwood2019turbulence} used the neural ODE approach to find an evolution equation for dissipation in decaying isotropic turbulence, showing that it outperformed a heuristic model. Neural ODEs have also been applied to flow around a cylinder in the time-periodic regime, where velocity field simulation data were projected onto 8 PCA modes, and the time evolution of these modes was determined \cite{rojas2021reducedorder}.

The present work combines nonlinear dimension reduction using autoencoders with the neural ODE method to model the dynamics of a system displaying spatiotemporal chaos, the Kuramoto-Sivashinsky equation, over a range of parameter values.  In Section \ref{sec:Framework}, we introduce the framework of our methodology. 
Section \ref{sec:autoencoders} briefly describes the results for the dimension reduction problem alone, then Section \ref{sec:dimredev} uses the reduced-dimensional descriptions to illustrate performance of the neural ODE approximation for closely spaced data.  An important conclusion here is that dimension reduction can improve neural ODE performance relative to predictions in the ambient space, where artifacts arise. Section \ref{sec:dataspacing} examines the role of data spacing on neural ODE performance, showing that even for  chaotic systems, widely spaced data can be used, within a fairly well-defined limit.  Finally, Section \ref{sec:dimdependence} shows comparisons of neural ODE performance with various degrees of dimension reduction, finding a ``sweet spot" in terms of performance vs.\ dimension. We summarize in Section \ref{sec:Conclusion}.
\section{Framework} \label{sec:Framework}
 

We consider data $u\in\reals^d$ that lies on a $d_\IM$-dimensional manifold $\mathcal{M}$ that is embedded in the ambient space $\mathbb{R}^d$ of the data. With the data, we find a coordinate transformation $\chi : \mathbb{R}^d \rightarrow \mathbb{R}^{d_\mathcal{M}}$ giving the state in the manifold coordinates, $h\in\mathbb{R}^{d_\mathcal{M}}$. We also find the inverse $\check{\chi} :\mathbb{R}^{d_\mathcal{M}} \rightarrow \mathbb{R}^d$ to reconstruct $u$ from $h$. Then we describe the dynamics on $\IM$ with the differential equation
	\begin{equation}\label{eq:ODE_red}
		\dfrac{dh}{dt}=g(h).
	\end{equation}
If we do not do any dimension reduction then Eq.\ \ref{eq:ODE_red} is the same as Eq.\ \ref{eq:ODE}. Of course, for an arbitrary data set, we do not know $d_\IM$ a priori, so in Section \ref{sec:dimdependence} we present results with various choices for $d_h$, where $h\in\mathbb{R}^{d_h}$. Using the mapping to the manifold coordinates, the evolution in the manifold coordinates, and the mapping back, we evolve new initial conditions forward in time. So, our task is to approximate $\chi$, $\check{\chi}$, and $g$. We also consider the case with no dimension reduction, where we determine $f(u)$, the right hand side (RHS) of the ODE in the ambient space. In general, there can be no expectation that any of these functions have a simple (e.g.\ polynomial) form, so here we approximate them with NNs, as detailed below.  

\begin{figure*}
	\centering
	\captionsetup[subfigure]{labelformat=empty}
	\begin{subfigure}[b]{17.2 cm}
		\includegraphics[trim=0 0 0 0,width=\textwidth,clip]{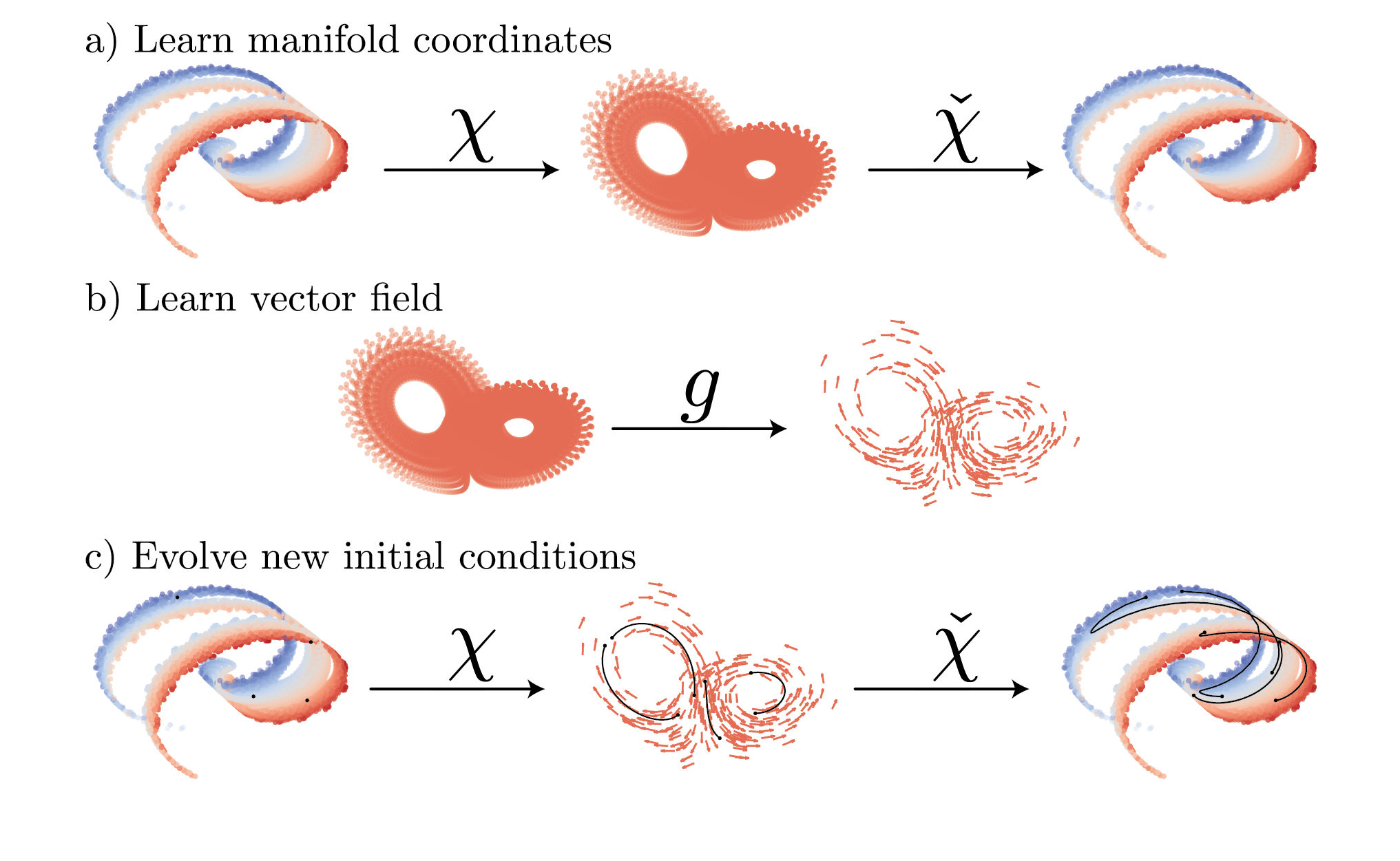}
		\caption{}
		\vspace{-10mm}
		\label{fig:Frameworka}
	\end{subfigure}
	\begin{subfigure}[b]{0\textwidth}\caption{}\vspace{-10mm}\label{fig:Frameworkb}\end{subfigure}
	\begin{subfigure}[b]{0\textwidth}\caption{}\vspace{-10mm}\label{fig:Frameworkc}\end{subfigure}
	\captionsetup{justification=raggedright}
	\vspace{-10mm}
	\caption{(a) Learning the three-dimensional manifold coordinates of the four-dimensional Lorenz butterfly wrapped on the Archimedean spiral (color is the fourth dimension). (b) Learning the vector field in the manifold coordinates. (c) Transforming new initial conditions (black dots) into the manifold coordinates, evolving them according to the learned vector field g (black curves), and transforming back into the original space.}
	\label{fig:Framework}
\end{figure*}

Figure \ref{fig:Framework} illustrates this framework on data from a solution of the Lorenz equation that we embedded in four dimensions by mapping the $z$ coordinate of the Lorenz equation to the Archimedean spiral. The change of coordinates for this embedding is given by $[u_1,u_2,u_3,u_4]=[x,y,\alpha z\cos \alpha z,\alpha z\sin \alpha z]$ where $x,y,z$ are the standard variables in the Lorenz equation and $\alpha=0.02$. In Fig.\ \ref{fig:Frameworka} we show an example of learning the manifold coordinates for this system. The three spatial dimensions for the embedded data are $u_1$, $u_3$, and $u_4$ and the color is $u_2$. For a trajectory to be chaotic it must lie on at least a three-dimensional manifold, so the minimum embedding dimension for a chaotic trajectory that requires all the steps in this framework is  four dimensions. Figure \ref{fig:Frameworkb} illustrates learning the vector field $g$ in the manifold coordinates. In Fig.\ \ref{fig:Frameworkc} we show how, after learning these functions, new initial conditions can be mapped to the manifold coordinates, evolved forward in time, and then mapped back to the ambient space.

To find $\chi$ and $\check{\chi}$, we represent them as NNs and find parameters that minimize a reconstruction loss averaged over a batch of data given by 
	\begin{equation}\label{eq:Auto}
		L=\left<||u-\check{\chi}(\chi(u;\theta_1);\theta_2)||^2\right>,
	\end{equation}
where $\theta_1$ and $\theta_2$ are the weights of $\chi$ and $\check{\chi}$, respectively. Here, and elsewhere, we use stochastic gradient descent methods to determine the parameters. Further details are given in Section \ref{sec:autoencoders}.
	This architecture is known as an undercomplete autoencoder, where $\chi$ is an encoder and $\check{\chi}$ is a decoder \cite{IanGoodfellowYoshuaBengio2017}. Autoencoders are widely used for dimension reduction; in the fluid mechanics context they have been used for many examples including flow around an airfoil \cite{Omata2019}, flow around a flat plate \cite{Nair2020}, Kolmogorov flow \cite{Page2020}, and channel flow \cite{Milano2002,Fukami2020}.



As noted above, to approximate $g$ in Eq.\ \ref{eq:ODE_red}, we use the neural ODE approach of Chen et al.\ \cite{Chen2019}.  
In the neural ODE framework we use $g$ to estimate $\tilde{h}(t_i+\tau)$, an approximation of the reduced state $h(t_i+\tau)$, at some time $t_i+\tau$ by integrating 
\begin{equation}\label{eq:ODENet_Int}
	\tilde{h}(t_i+\tau)=h(t_i)+\int_{t_i}^{t_i+\tau}g(h(t);\theta_3) dt,
\end{equation}
where $\theta_3$ is the set of weights of the NN. Then $g$ is learned by minimizing the difference between the prediction of the state ($\tilde{h}(t_i+\tau)$) and the known state ($h(t_i+\tau)$), in the manifold coordinates, at that time. Specifically, the loss we minimize is
\begin{equation}\label{eq:ODENet}
	J=\left<||h(t_i+\tau)-\tilde{h}(t_i+\tau)||_1\right>.
\end{equation}
Here $||.||_1$ denotes the $L_1$-norm -- of course other norms can be used.
By taking this approach we can estimate $g$ from data spaced further apart in time than when $g$ is estimated directly from finite differences.
%
%
%
 The difficulty comes in determining the gradient of the loss with respect to the weights of the NN, $\partial J/\partial \theta_3$. 

One approach is to use automatic differentiation to back-propagate through all of the steps of the time integration scheme used to solve \ref{eq:ODENet_Int}. Another approach involves solving an adjoint problem backwards in time. A drawback of back-propagating through the solver is that all of the data must be stored at each time-step to calculate the gradient, which can lead to memory issues \cite{Chen2019}. However, we consider a chaotic system which puts an implicit constraint on how far apart data can feasibly be sampled and still yield a good estimate of $g$. We reach this limit before the memory limit of back-propagation becomes an issue. In our trials, the adjoint method yielded results in good agreement with back-propagation, but required longer computation times for training, so we chose to use the back-propagation approach.



The data sets we consider are numerical solutions of the 1D Kuramoto-Sivashinsky equation (KSE),
\begin{equation}\label{eq:KSE}
	\dfrac{\partial v}{\partial t}=-v\dfrac{\partial v}{\partial x}-\dfrac{\partial^2 v}{\partial x^2}-\dfrac{\partial^4 v}{\partial x^4}, 
\end{equation}
in a domain of length $L$ with periodic boundary conditions. The domain size determines the types of dynamics this system exhibits. For the domain sizes we consider, trajectories exhibit sustained chaos. The dimension of the manifold that contains the global attractor has been computationally approximated using several approaches \cite{Yang2009,Ding2016,Linot2020}. We generate high-dimensional state representations ($u\in\mathbb{R}^d$) via a Galerkin projection  of $v$ onto Fourier modes. Then, we use an exponential time differencing method \cite{Kassam2005} to integrate the resulting system of ordinary differential equations forward in time. The code used is available from Cvitanovi\'c et al.\ \cite{ChaosBook}. The data vectors $u$ that we use are solutions $v$ of the KSE on $d=64$ equally-spaced grid points in the domain.

\section{Results} \label{sec:Results}

Section \ref{sec:autoencoders} briefly describes the dimension reduction (manifold representation results).  Section \ref{sec:dimredev} considers neural ODE predictions, with and without dimension reduction, for closely spaced data. Section \ref{sec:dataspacing} shows the effect of data spacing on the results, and Section \ref{sec:dimdependence} illustrates the effect of the degree of dimension reduction. We summarize in Section \ref{sec:Conclusion}.


We consider datasets for three domain sizes, $L=22$, $44$, and $66$, where we estimate the manifold dimension to be $d_\mathcal{M}=8$, $18$, and $28$, as explained in the following section. The relevant timescale for each of these system is the Lyapunov timescale (inverse of the largest Lyapunov exponent), which we previously found to be be very close to the integral time for KSE data \cite{Linot2020}. For these domain sizes, the Lyapunov times are $\tau_L=22.2$, $12.3$, and $11.6$, respectively \cite{Ding2016,Edson2019}. We use $10^5$ time units of data for training at each domain size, and vary how frequently we sample this dataset in time. 
For training the NNs we use Keras \cite{chollet2015keras} for the autoencoders and PyTorch \cite{Paszke2019} for the neural ODEs. The neural ODE code is a modification on the code used in \cite{Chen2019}.

\subsection{Dimension reduction with autoencoders}\label{sec:autoencoders}

In each section we use autoencoders to approximate the map to the manifold coordinates $\chi$ and back $\check{\chi}$. We found in \cite{Linot2020} that a useful way to represent the map to the manifold coordinates ($h=\chi(u)$) is as a difference from a linear projection of the data onto the leading $d_h$ PCA modes:
	\begin{equation} \label{eq:hidden}
		h=E(U^T u;\theta_1)+P_{d_h}U^T u.
	\end{equation}
Here $E$ is a NN, $U$ is the full PCA matrix, and $P_{d_h}$ is the projection onto the leading $d_h$ modes. Similarly, we can learn a difference for the decoder ($\tilde{u}=\check{\chi}(h)$)
\begin{equation} \label{eq:Autoencoder}
	\tilde{u}=UD(h;\theta_2)+U\begin{bmatrix}
		{h} \\ 0
	\end{bmatrix},
\end{equation}
where $D$ is a NN.
By taking this approach we simplify the problem such that the NN only needs to correct upon PCA. We refer to this as a hybrid NN (HNN). 

\begin{figure} 
	\includegraphics[trim=0 0 0 0,width=8.6 cm,clip]{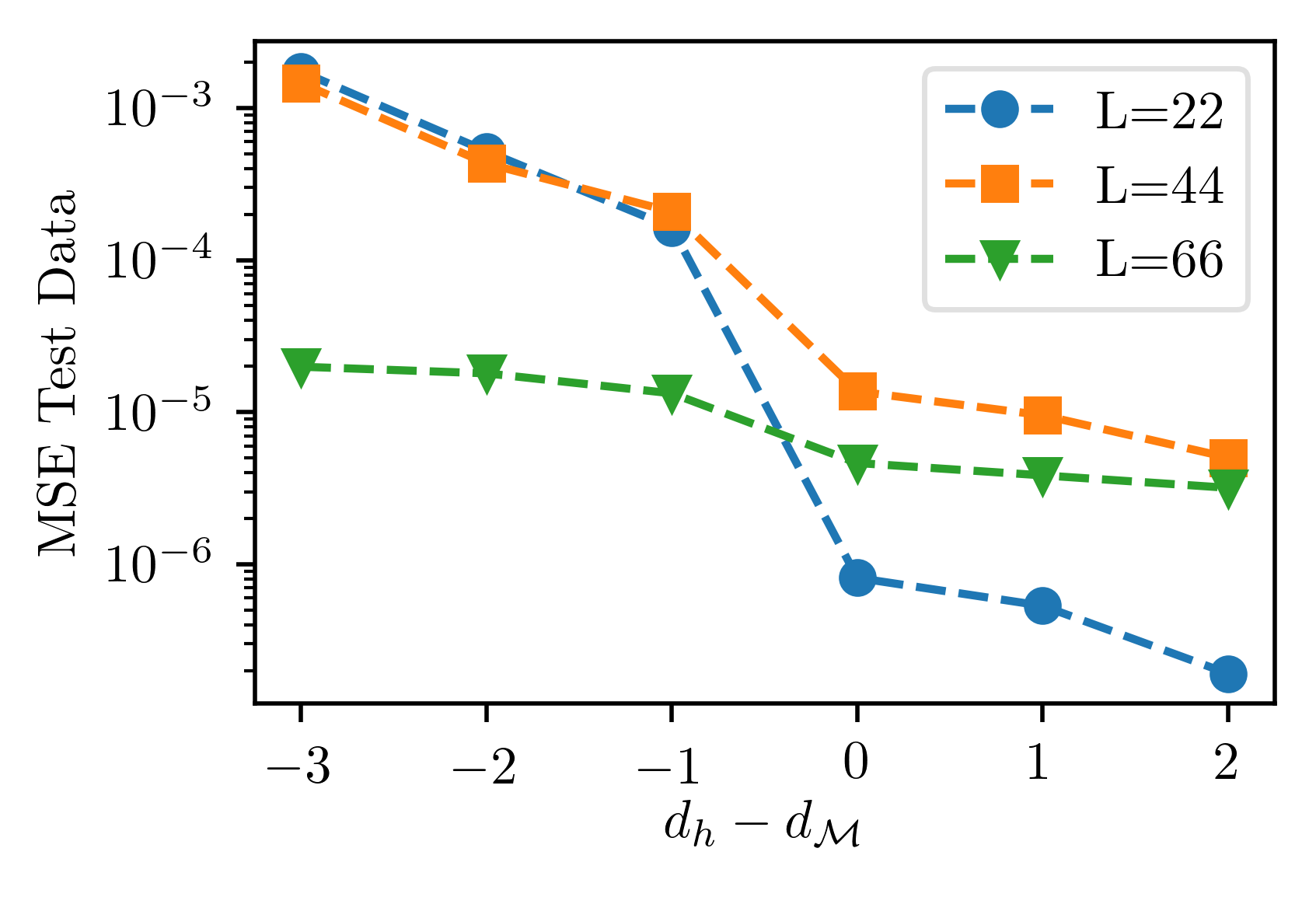}
	\caption{MSE of test data as a function of dimension for domain sizes $L=22$, $44$, and $66$. The expected manifold dimensions are $d_\IM=8$, $18$, and $28$ for these domain sizes, respectively.}
	\label{fig:Dimension}
\end{figure} 

In \cite{Linot2020} we trained these HNN using an Adam optimizer with a learning rate of 0.001 at multiple dimensions. Figure \ref{fig:Dimension} shows the mean squared reconstruction error (MSE) of a test data set not used in training at various dimensions for each of the domain sizes. When using this hybrid approach we see a significant drop in the MSE at dimensions of $d_\mathcal{M}=8$, $18$, and $28$ for the domain sizes $L=22$, $44$, and $66$, respectively. The observed approximately linear relationship between dimension and domain size is in agreement with the scaling of physical modes in \cite{Yang2009}, and the dimension for $L=22$ is in agreement with estimates by Ding et al.\ \cite{Ding2016}.

These results guide our selection of dimension in the next two sections. Furthermore, we find that at the manifold dimension that the same MSE is achieved whether we use the full HNN or simply project on to $d_\IM$ PCA modes ($E(U\trans u)=0$). This suggests, for the KSE with periodic boundary conditions, that the leading PCA coefficients fully parameterize the state. So, in the next two sections, $\chi$ is the projection onto $d_\IM$ PCA modes and $\check{\chi}$ is a NN that approximates the remaining PCA coefficients from the leading ones. Table \ref{Table} shows the architectures for the autoencoders and for the NNs used for time evolution in Sections \ref{sec:dimredev} and \ref{sec:dataspacing}.

\subsection{Effect of Dimension Reduction on Time Evolution}\label{sec:dimredev}

Now that we have a map to the manifold coordinates and back, the next step is approximating the ODE. Here we train three different types of ODEs to evaluate the impact of mapping the data to the manifold coordinates. We learn the RHS of the ODE in the manifold coordinates: $\dot{h}=g(h)$, in physical space: $\dot{u}=f(u)$, and in the space of the spatial Fourier coefficients: $\dot{\hat{u}}=\hat{f}(\hat{u})$, where $\hat{u}=\mathcal{F}(u)$ and $\mathcal{F}$ is the discrete Fourier transform. For the last two cases there is no dimension reduction, so $\chi$ is the identity in the first case and the discrete Fourier transform in the second, with $\check{\chi}$ being, again, the identity in the first case and the discrete inverse Fourier transform in the second.


\begin{table}
	\captionsetup{justification=raggedright}
	\caption{Architectures of NNs and matrices used in Sections \ref{sec:autoencoders}-\ref{sec:dataspacing}. ``Shape" indicates the dimension of each layer, and ``activation" the corresponding activation functions (S is the sigmoid activation) \cite{IanGoodfellowYoshuaBengio2017}.}
	\resizebox{.65\textwidth}{!}{%
		\begin{tabular}{l*{6}{c}r}
			              & Function & Shape & Activation \\
			\hline
			Encoder $L=22,44,66$	& $\chi$ 			& $d:d_\mathcal{M}$ & linear  \\
			Decoder $L=22,44$	& $\check{\chi}$ 	& $d_\mathcal{M}:500:d$  & S:linear  \\
			Decoder $L=66$		& $\check{\chi}$ 	& $d_\mathcal{M}:500:500:d$  & S:S:linear  \\
			ODE					& $f/g$ 		& $d/d_\mathcal{M}:200:200:200:d/d_\mathcal{M}$ & S:S:S:linear  \\
			Discrete Map 		& $G$ 				& $d_\mathcal{M}:200:200:200:d_\mathcal{M}$ & S:S:S:linear  \\
		\label{Table}
		\end{tabular}}
\end{table}

We train each of the NN with data spaced 0.25 time units in these trials. This timescale is substantially shorter than the Lyapunov time, which decouples the issue of dimension from that of time resolution. Each NN is trained until the loss levels off using an Adam optimizer with a learning rate of $10^{-3}$ that we drop to $10^{-4}$ halfway through training. First the autoencoder is trained, then the neural ODE. To avoid spurious results, we train 5 autoencoders then 10 ODEs, and select the combination of autoencoder and ODE that provides the best short-time tracking. Due to random weight initialization and the stochastic nature of the optimization, we train multiple models and use the one that performs best. 

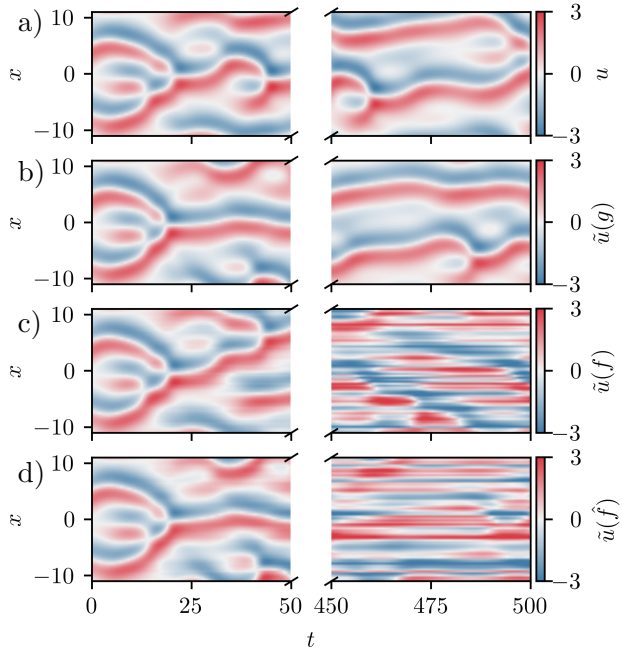
\begin{figure}
	\centering
	\captionsetup[subfigure]{labelformat=empty}
	\begin{subfigure}[b]{8.6 cm}
		\includegraphics[trim=0 0 0 0,width=\textwidth,clip]{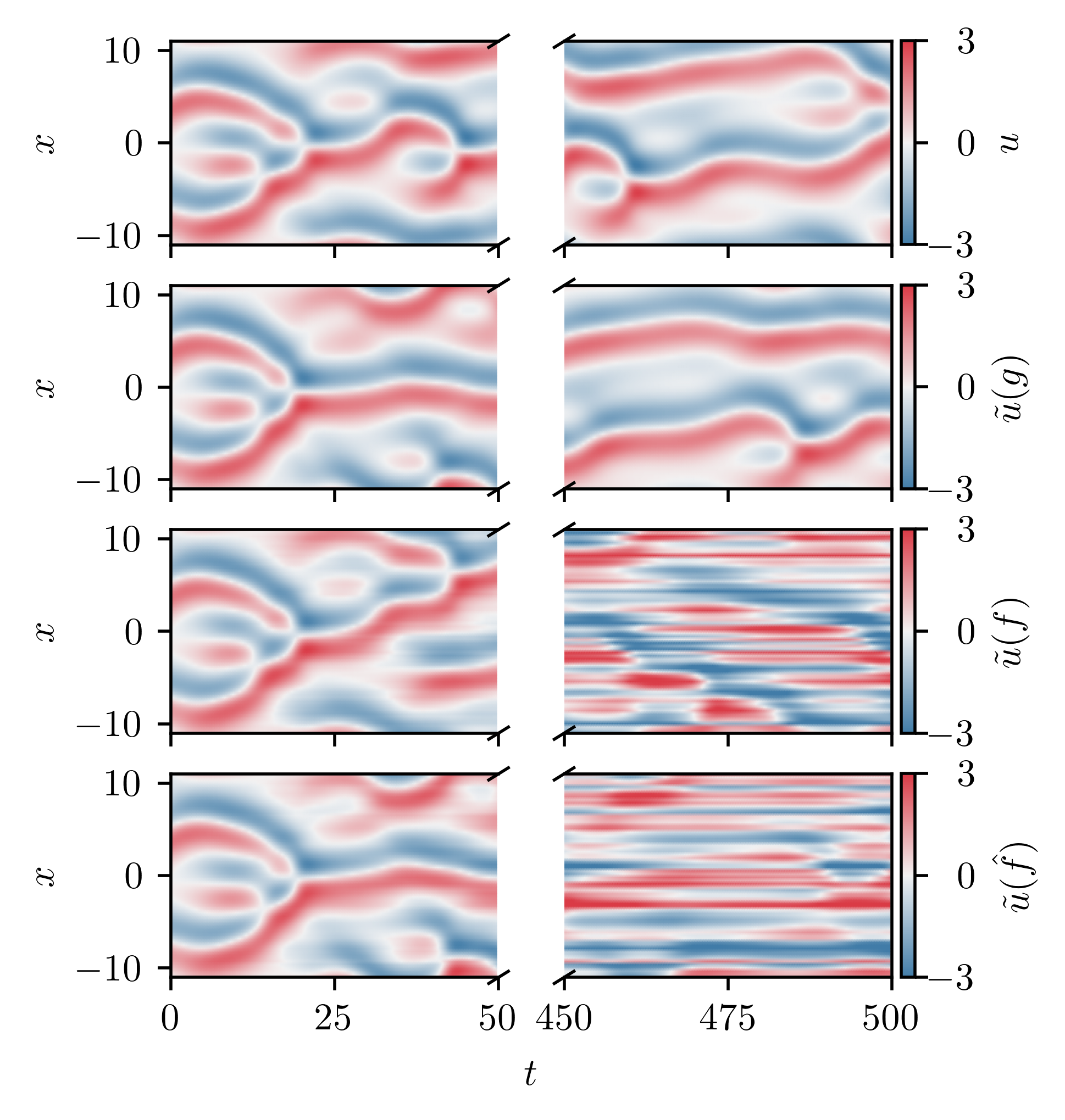}
		\begin{picture}(0,0)
		\put(-115,263){\contour{white}{ \textcolor{black}{a)}}}
		\put(-115,206){\contour{white}{ \textcolor{black}{b)}}}
		\put(-115,149){\contour{white}{ \textcolor{black}{c)}}}
		\put(-115,92){\contour{white}{ \textcolor{black}{d)}}}
		\end{picture}
		\caption{}
		\vspace{-10mm}
		\label{fig:Traja}
	\end{subfigure}
	\begin{subfigure}[b]{0\textwidth}\caption{}\vspace{-10mm}\label{fig:Trajb}\end{subfigure}\begin{subfigure}[b]{0\textwidth}\caption{}\vspace{-10mm}\label{fig:Trajc}\end{subfigure}\begin{subfigure}[b]{0\textwidth}\caption{}\vspace{-10mm}\label{fig:Trajd}\end{subfigure}    
	\vspace{-1.5\baselineskip}
	\captionsetup{justification=raggedright}
	\caption{Examples of trajectory predictions for different models. (a) is the true trajectory, (b) is the predicted trajectory when approximating $g$ for $d_\mathcal{M}=8$. (c) is the predicted trajectory when approximating $f$. (d) is the predicted trajectory when approximating $\mathcal{F}(f)$.} 
	\label{fig:Fullspace} 
	\vspace{-5mm}
\end{figure}

In Fig.\ \ref{fig:Fullspace} we compare the short and long-time trajectories between the data at $L=22$ (Fig.\ \ref{fig:Traja}) and the NN models (Fig.\ \ref{fig:Trajb}-\ref{fig:Trajd}). The same initial condition, which is on the attractor, is used for all cases. The first model prediction, shown in Fig.\ \ref{fig:Trajb}, comes from $g$ (the low-dimensional ODE), the second model prediction, shown in Fig.\ \ref{fig:Trajc}, comes from $f$ (the full physical space ODE), and the third model prediction, shown in Fig.\ \ref{fig:Trajd}, comes from $\hat{f}$ (the full Fourier space ODE). 
For each of these figures, the first 50 time units are shown on the left and 450-500 time units are shown on the right. 

For this domain size, the Lyapunov time is $\tau_L=22.2$. All three of the models are able to generate good predictions for times up to about $t=30$; this is a reasonable quantitative prediction horizon for a data-driven model of chaotic dynamics. At longer times, a good model should still yield behavior consistent with the data, and indeed the reduced model, Fig.\ \ref{fig:Trajb}, no longer matches the true solution, but continues to exhibit realistic dynamics. In the next section we discuss the long-time behavior in more detail. However, the full state model predictions in Figs. \ref{fig:Trajc} and \ref{fig:Trajd} develop high wavenumber striations at long times, and are thus not faithful to the dynamics. Training high-dimensional neural ODEs is computationally more expensive, and the predictions from these models are worse.


\begin{figure*}
	\centering
	\captionsetup[subfigure]{labelformat=empty}
	\begin{subfigure}[b]{17.2 cm}
		\includegraphics[trim=0 0 0 0,width=\textwidth,clip]{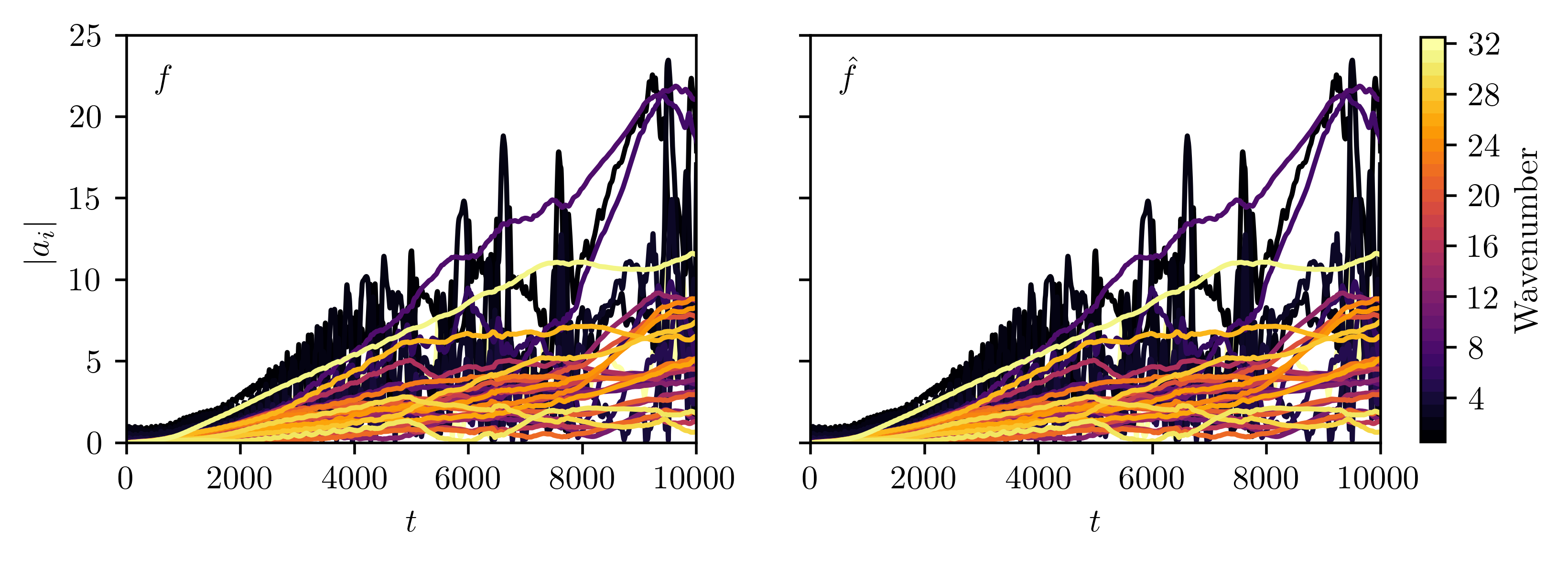}
		\begin{picture}(0,0)
		\put(-240,170){a)}
		\put(-10,170){b)}
		\end{picture}
		\caption{}
		\vspace{-10mm}
		\label{fig:Wavenumbers_Real}
	\end{subfigure}
	\begin{subfigure}[b]{0\textwidth}\caption{}\vspace{-10mm}\label{fig:Wavenumbers}\end{subfigure}
	\captionsetup{justification=raggedright}
	\caption{Magnitude of Fourier modes over time when evolving an initial condition with a neural ODE approximation of (a) the full-space dynamics in real space ($f$) and of (b) the full-space dynamics in Fourier space ($\hat{f}$) for $L=22$.}
	\label{fig:Wavenumbers_Both}
	\vspace{-5mm}
\end{figure*}
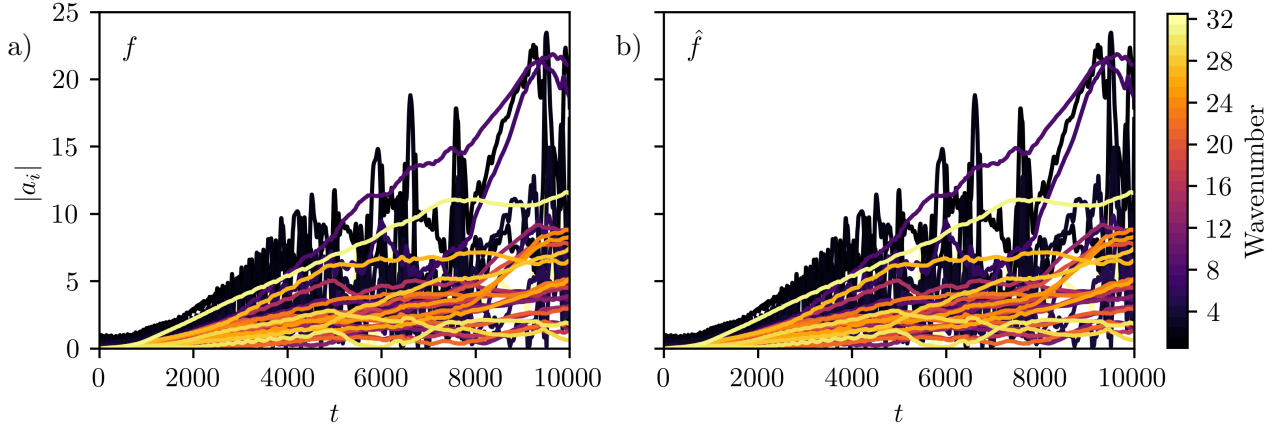

To understand this high wavenumber behavior, we plot the magnitude of the Fourier modes as a function of time in Fig.\ \ref{fig:Wavenumbers_Both}. In both cases, there is a linear growth in the high wavenumber modes despite the sigmoid activation functions used in the NN being bounded. This appears to arise because perturbations off the attractor cause the sigmoids to go to either 0 or 1 resulting in a constant bias term on the RHS of the neural ODE. However, this phenomena is not exclusive to our choice of sigmoid activations. When we use other activation functions, like hyperbolic tangent and rectified linear units, we also see a bias term on the RHS.

 

In the true system these high wavenumbers are strongly damped due to hyperdiffusivity in the KSE, so the data used to train these models contains little content in these wavenumbers. Furthermore, the prediction horizon in the training of the neural ODE is short  (0.25 here), so small errors in high wavenumbers have a negligible effect on the loss. 
The combination of these factors is why the trajectories of the high-dimensional models leave the attractor, while the trajectories of the low-dimensional model do not. This same behavior appears when modeling the full state for the larger domain sizes $L=44$ and $66$.

%

\subsection{Effect of Temporal Data Spacing on Time Evolution Prediction}\label{sec:dataspacing}

\begin{figure}
	\centering
	\captionsetup[subfigure]{labelformat=empty}
	\begin{subfigure}[b]{8.6 cm}
		\includegraphics[trim=0 0 0 0,width=\textwidth,clip]{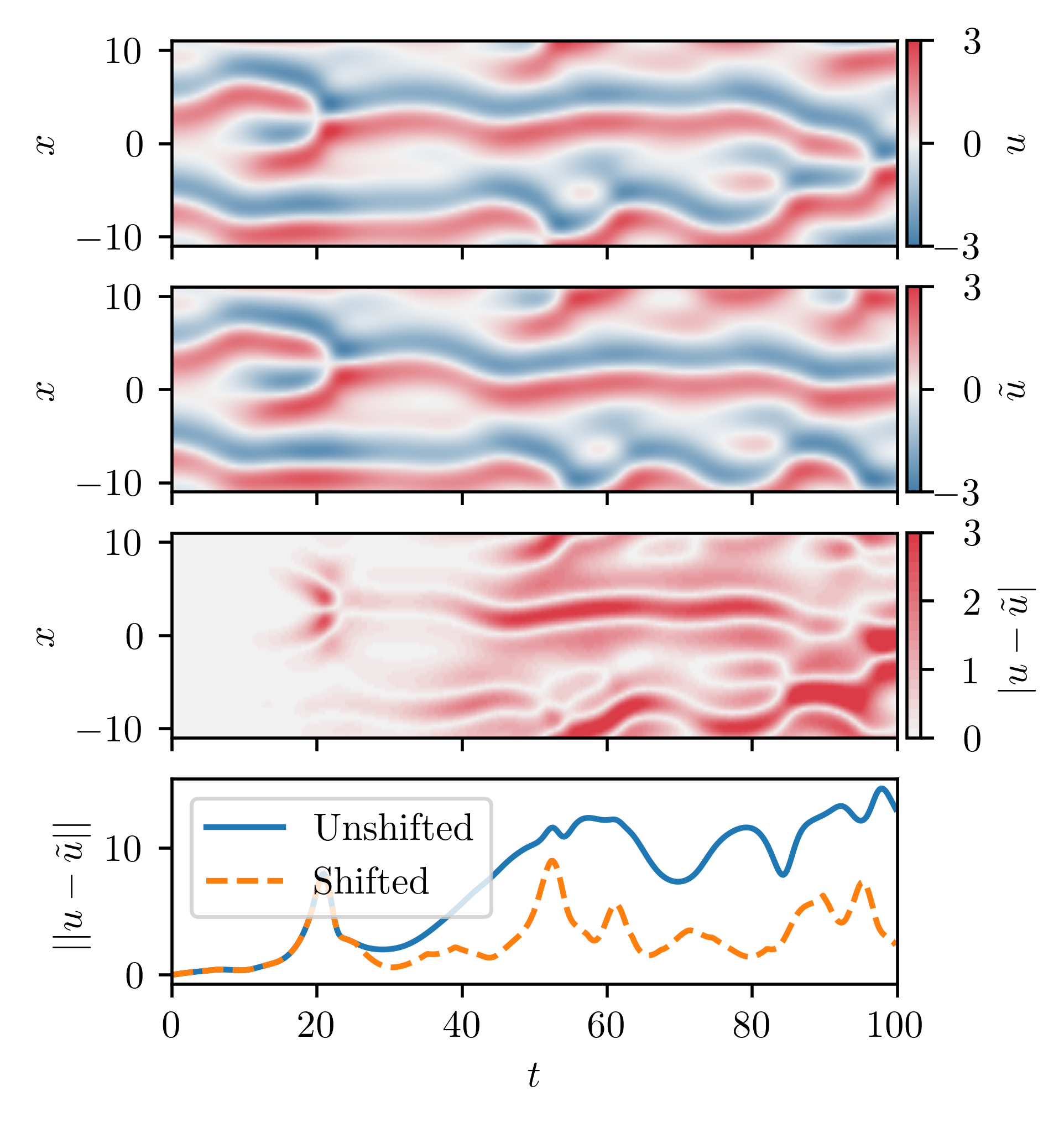}
		\begin{picture}(0,0)
		\put(-115,263){\contour{white}{ \textcolor{black}{a)}}}
		\put(-115,206){\contour{white}{ \textcolor{black}{b)}}}
		\put(-115,149){\contour{white}{ \textcolor{black}{c)}}}
		\put(-115,96){\contour{white}{ \textcolor{black}{d)}}}
		\end{picture}
		\caption{}
		\vspace{-10mm}
		\label{fig:Exa}
	\end{subfigure}
	\begin{subfigure}[b]{0\textwidth}\caption{}\vspace{-10mm}\label{fig:Exb}\end{subfigure}\begin{subfigure}[b]{0\textwidth}\caption{}\vspace{-10mm}\label{fig:Exc}\end{subfigure}\begin{subfigure}[b]{0\textwidth}\caption{}\vspace{-10mm}\label{fig:Exd}\end{subfigure}    
	\vspace{-1.5\baselineskip}
	\captionsetup{justification=raggedright}
	\caption{Example of trajectory and predictions for data spaced $\tau=10$ apart: (a) true trajectory; (b) predicted trajectory; (c) absolute error between the two trajectories; (d) norm of the error. The norm of the error is shown for the raw data, and for the spatial shift that minimizes this error at each time.}
	\label{fig:Example} 
	\vspace{-5mm}
\end{figure}
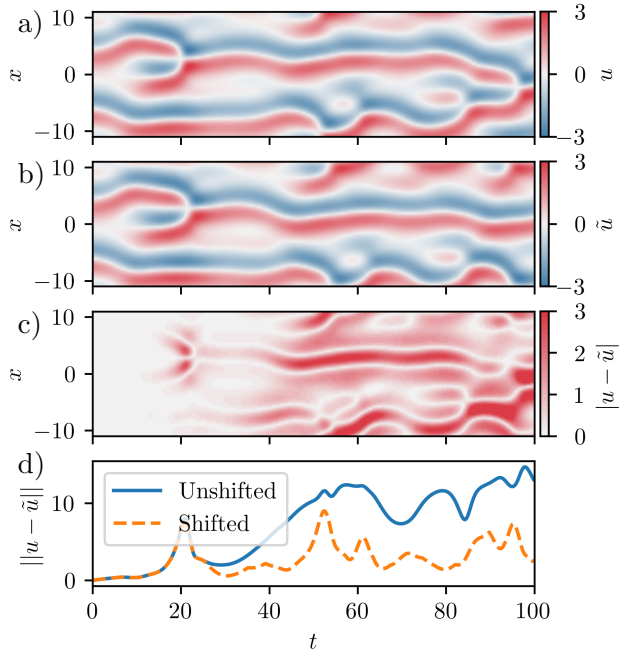

Now we consider only reduced-dimension ODE models, and address the dependence of model quality on the temporal spacing $\tau$ between data points. For $L=22$, $44$, and $66$, we select the manifold dimensions mentioned in Section \ref{sec:autoencoders} -- $d_\IM=8$, $18$, and $28$, respectively. We judge the model performance based on, first, short-time tracking for $L=22$ and then long-time statistical agreement for all domain sizes. In these cases autoencoders and ODEs were both trained with $\tau$ much larger than the value of 0.25 used in Section \ref{sec:dimredev}, while keeping all other training parameters the same (e.g. architecture, optimizer, dimension). Figure \ref{fig:Example} compares a true short-time trajectory (\ref{fig:Exa}) to the model prediction (with $\tau=10$) of this trajectory (\ref{fig:Exb}) starting from the same initial condition. Qualitatively the space-time plots exhibit similar behavior over around 80 time units. The magnitude of the difference is shown in \ref{fig:Exc}. Here we see the difference growing at around 40 time units due primarily to a growing spatial phase difference between the true and predicted results. Figure \ref{fig:Exd} shows the norm of this difference and the minimum value of the norm when shifting $\tilde{u}$ to any position in the domain. The minimum translated difference, in \ref{fig:Exd}, remains much smaller than the standard Euclidean distance indicating that, in this case, the phase difference causes most of the error.

We now consider ensemble average quantities to better understand the reconstruction quality of the models as a function of $\tau$. In addition to the ODE models, we also consider discrete-time maps $h(t_i+\tau)=G(h(t_i))$. 
 Details of the discrete-time map NN architecture is given in Table \ref{Table}. 

\begin{figure} 
	\includegraphics[trim=0 0 0 0,width=8.6 cm,clip]{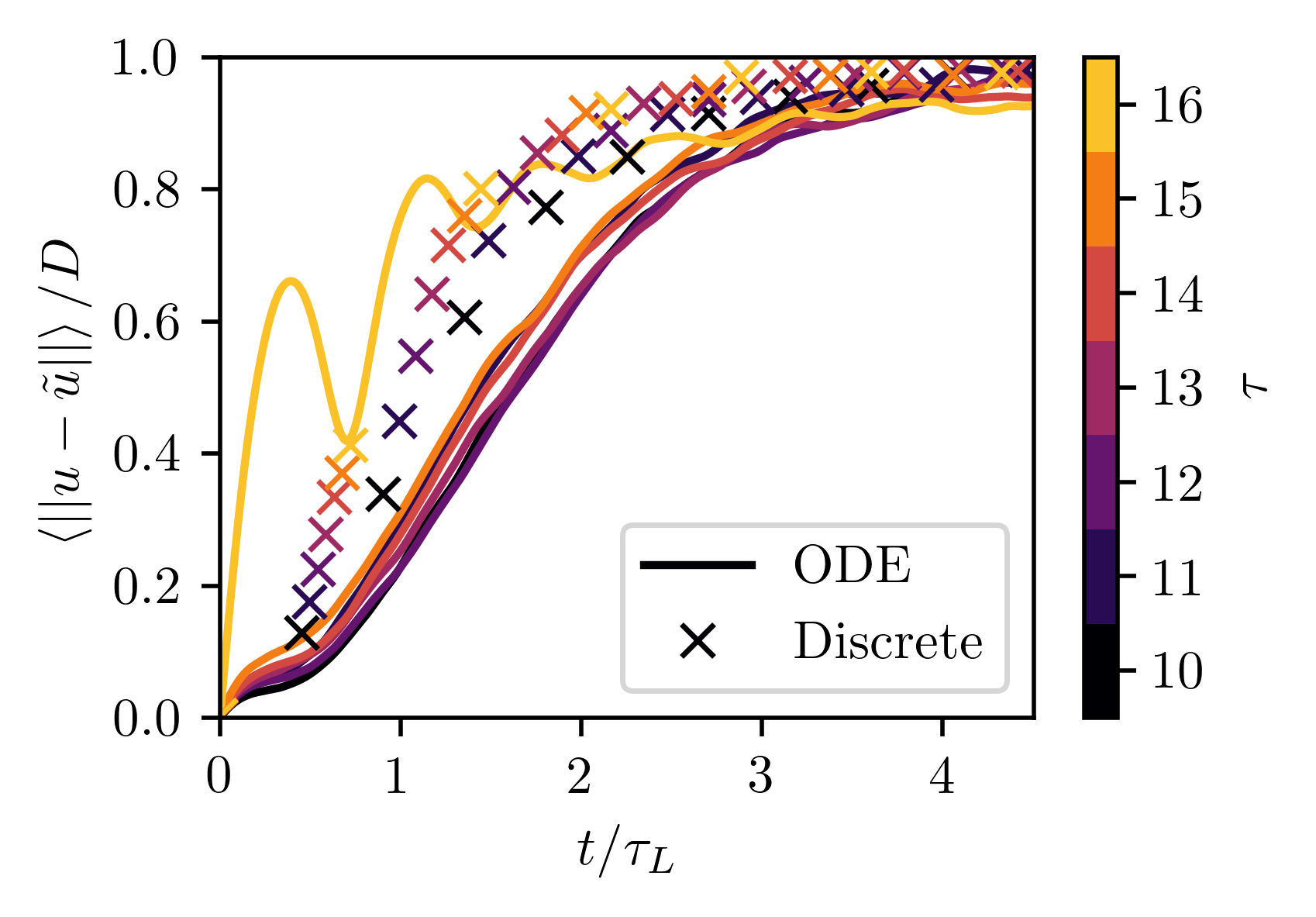}
	\caption{Normalized difference between trajectory and prediction as a function of time for different training data spacing.}
	\label{fig:Difference}
\end{figure} 

Figure \ref{fig:Difference} shows the root mean squared difference between the exact trajectory and ODE/discrete-time map models as a function of time, averaged over 1000 initial conditions and normalized by the root mean squared difference between random training data, which is denoted with $D$. In this figure, the ODE models and discrete time maps use data spaced $\tau=10-16$. For $\tau<10$, there is little improvement in the ODE models' performance.
ODE predictions at and below $\tau=15$ all track well for short times, with the error being halfway to random at around $t\sim 1.5\tau_L$, and diverge at long times, with the error leveling off at around $t\sim 3\tau_L$. 
Then, performance degrades sharply at $\tau=16$ (yellow curve). 
In all cases, the discrete-time map performs worse than the ODE with the same data spacing. Furthermore, although we do not show it here, the performance for discrete-time maps becomes even worse when trying to interpolate between prediction times. This is a significant drawback not seen when using ODEs.


\begin{figure} 
	\includegraphics[trim=0 0 0 0,width=8.6 cm,clip]{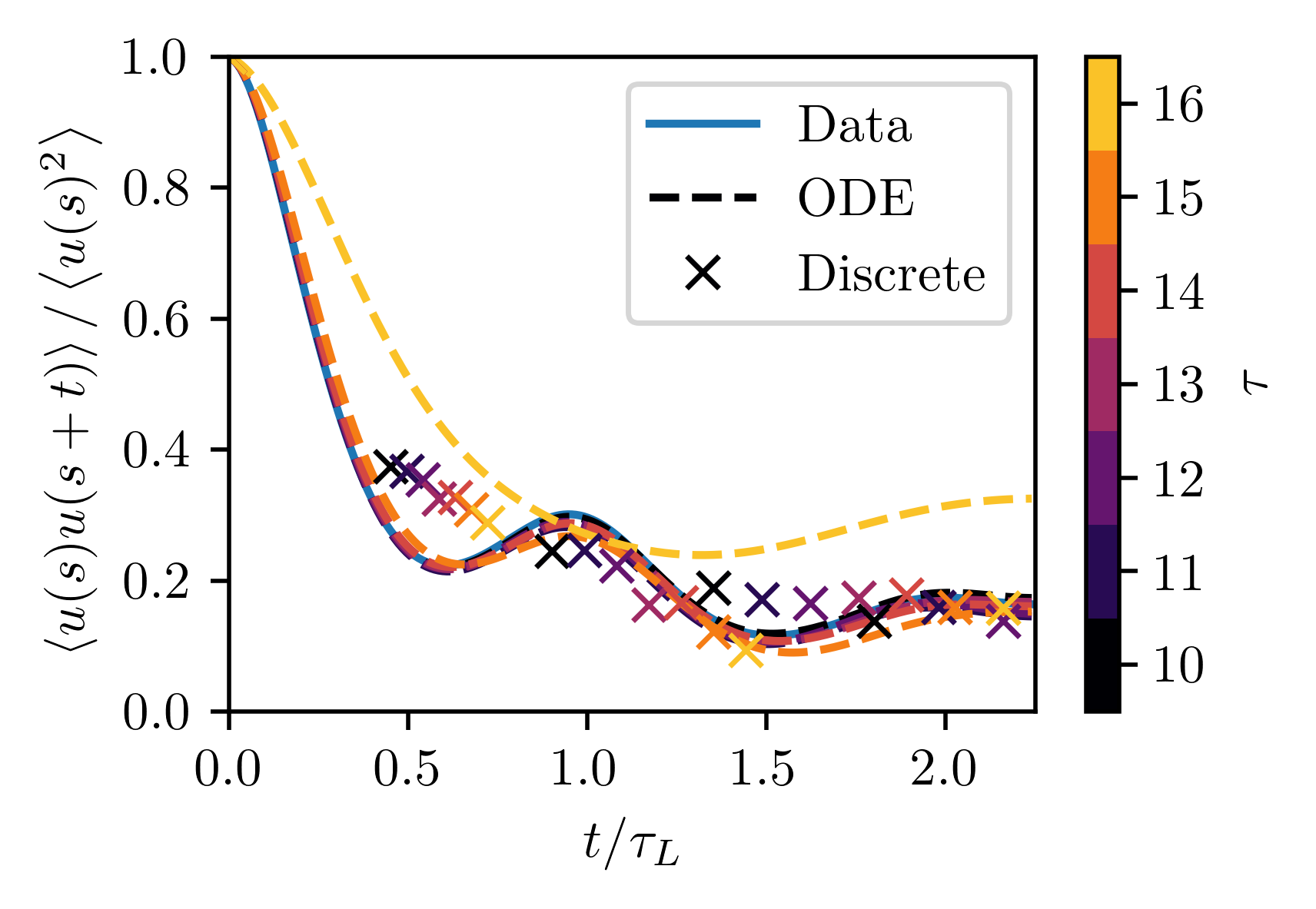}
	\caption{Temporal autocorrelation for models learned with different data spacing.}
	\label{fig:TempCor}
\end{figure}

A similar trend appears in the temporal autocorrelation. Figure \ref{fig:TempCor} shows the temporal autocorrelation of $u$ at a given gridpoint averaged over space and 1000 initial conditions.
The temporal autocorrelation of the neural ODE matches the exact temporal autocorrelation well for data spaced up to $\tau=15$. Then, there is an abrupt deviation from the true solution at $\tau=16$. 
Also, in Fig.\ \ref{fig:TempCor} we show the temporal autocorrelation for discrete-time maps. At $\tau=10$ discrete-time maps perform worse than all ODEs below $\tau=15$, and predictions worsen when increasing $\tau$.


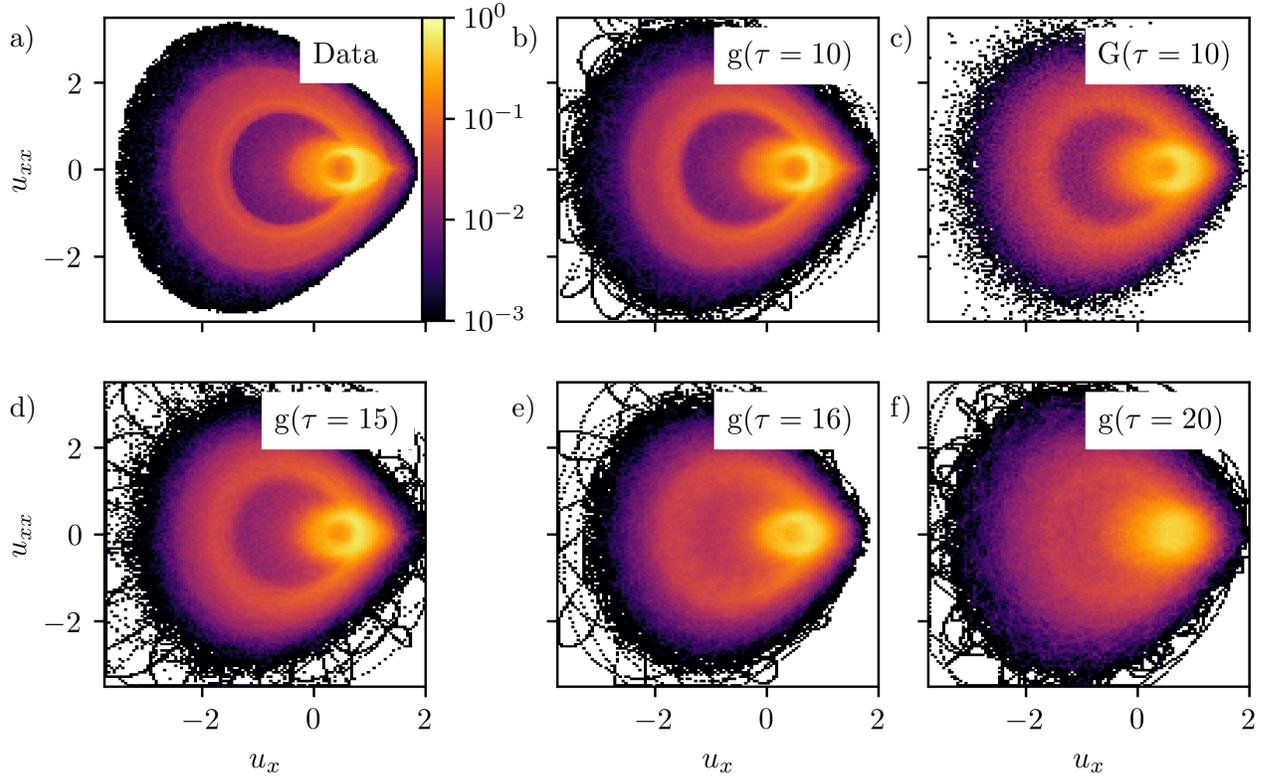
\begin{figure*}
	\centering
	\captionsetup[subfigure]{labelformat=empty}
	\begin{subfigure}[b]{17.2 cm}
		\includegraphics[trim=0 0 0 0,width=\textwidth,clip]{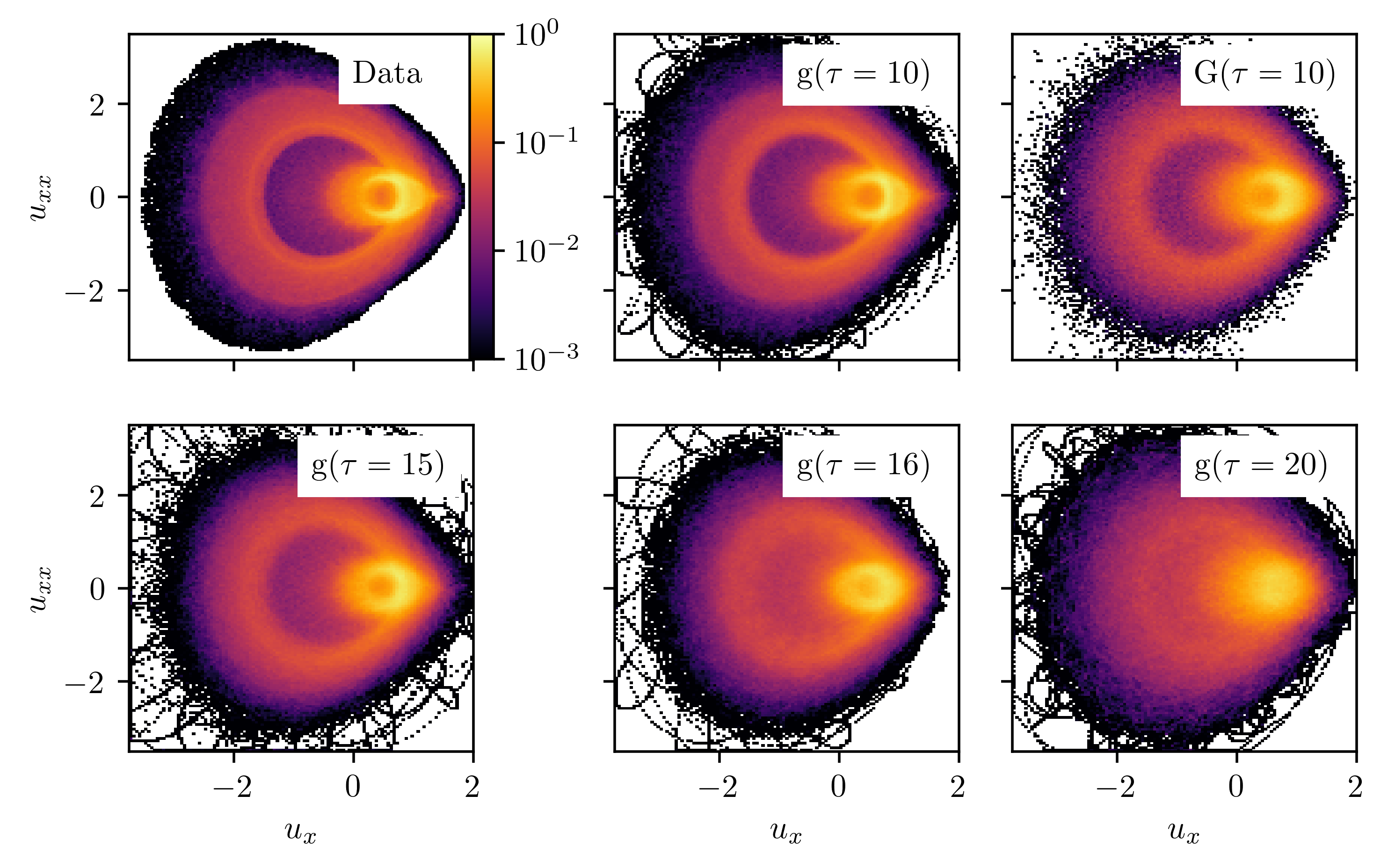}
		\begin{picture}(0,0)
		\put(-235,305){a)}
		\put(-45,305){b)}
		\put(98,305){c)}
		\put(-235,165){d)}
		\put(-45,165){e)}
		\put(98,165){f)}
		\end{picture}
		\caption{}
		\vspace{-10mm}
		\label{fig:PDF_True}
	\end{subfigure}
	\begin{subfigure}[b]{0\textwidth}\caption{}\vspace{-10mm}\label{fig:PDFg10}\end{subfigure}\begin{subfigure}[b]{0\textwidth}\caption{}\vspace{-10mm}\label{fig:PDFF10}\end{subfigure}\begin{subfigure}[b]{0\textwidth}\caption{}\vspace{-10mm}\label{fig:PDFg15}\end{subfigure}\begin{subfigure}[b]{0\textwidth}\caption{}\vspace{-10mm}\label{fig:PDFg16}\end{subfigure}\begin{subfigure}[b]{0\textwidth}\caption{}\vspace{-10mm}\label{fig:PDFg20}\end{subfigure}
	\captionsetup{justification=raggedright}
	\caption{(a) joint PDF at $L=22$. (b) and (c) joint PDFs when approximating the ODE and the discrete-time map, respectively, at $\tau=10$. (d)-(f) joint PDFs when approximating the ODE for $\tau=15$, $16$, and $20$.}
	\label{fig:PDF}
	\vspace{-5mm}
\end{figure*} 

The other half of evaluating a model is determining if trajectories stay on the attractor at long times. For this purpose, we consider the long-time joint PDF of the first ($u_x$) and second ($u_{xx}$) spatial derivatives of the solution. We select these quantities because both are relevant to the energy balance for the KSE ($\left<u_x^2\right>$ is the energy production and $\left<u_{xx}^2\right>$ the dissipation), and joint PDFs are more difficult to reconstruct than single PDFs. 
 In Fig.\ \ref{fig:PDF} the joint PDF for data is compared to various models. The colormap is a log scale, with low probability excursions in black, and no data in white regions. When $\tau=10$, the ODE model matches well, with differences primarily in the low probability excursions, while at $\tau=10$ the discrete-time mapping appears more diffuse in the high probability region and matches poorly. At $\tau=15$, the joint PDF for the ODE prediction still matches well, degrading once $\tau\geq 16$. This deterioration becomes more evident at $\tau=20$.


\begin{figure} 
	\includegraphics[trim=0 0 0 0,width=8.6 cm,clip]{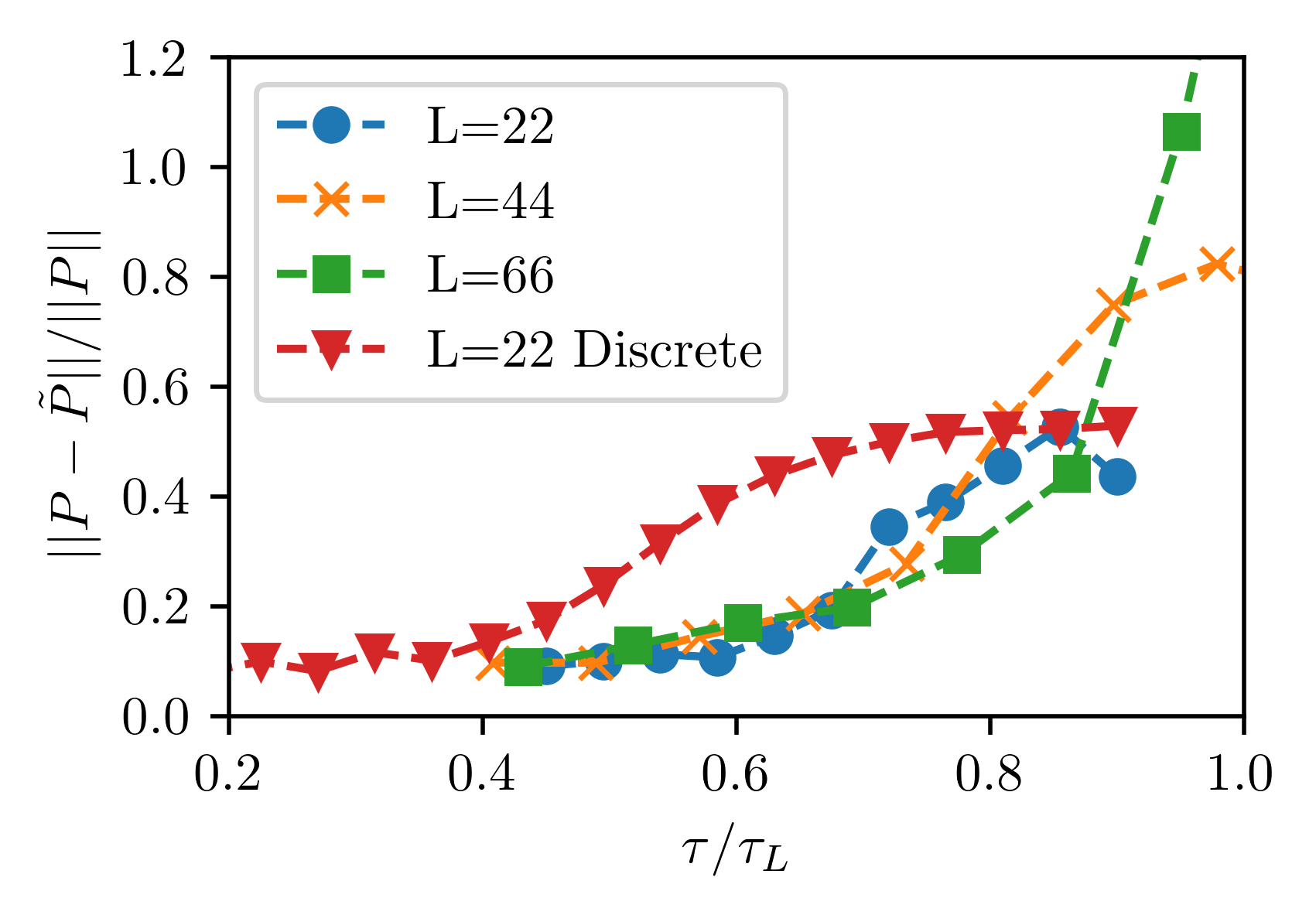}
	\caption{Relative error in Joint PDFs for models trained with different data spacing. Here time is scaled with Lyapunov time for the corresponding domain size.}
	\label{fig:JointPDFError_Temp}
\end{figure}

We quantify the difference between true and predicted PDFs by considering the relative error $||P-\tilde{P}||/||P||$ where $P$ corresponds to the true joint PDF and $\tilde{P}$ corresponds to the prediction. This relative error appears in Fig.\ \ref{fig:JointPDFError_Temp} for various $\tau$, where we normalize by the Lyapunov time $\tau_L$. For all domain sizes, the joint PDF reconstruction error is small for neural ODE models when data is spaced below $\sim$0.7 Lyapunov times. In contrast, good reconstruction of the joint PDF with the discrete-time model, for $L=22$, requires $\tau/\tau_L\lesssim 0.4$.

\subsection{Effect of Model Dimension on Time Evolution Predictions}\label{sec:dimdependence}

\begin{table}
	\captionsetup{justification=raggedright}
	\caption{Architectures of NNs used in Section \ref{sec:dimdependence}. Labels are the same as in Table \ref{Table}.}
	\resizebox{.65\textwidth}{!}{%
		\begin{tabular}{l*{6}{c}r}
			              & Function & Shape & Activation \\
			\hline
			Encoder $L=22$		& $\chi$ 			& $d:500:d_h$  & S:linear  \\
			Encoder	$L=44,66$	& $\chi$ 			& $d:500:500:d_h$  & S:S:linear  \\
			Decoder $L=22$		& $\check{\chi}$ 	& $d_h:500:d$  & S:linear  \\
			Decoder $L=44,66$	& $\check{\chi}$ 	& $d_h:500:500:d$  & S:S:linear  \\
			ODE					& $g$ 				& $d_h:200:200:200:d_h$ & S:S:S:linear  \\
		\label{Table2}
		\end{tabular}}
\end{table}

In the previous sections, the model dimension was fixed at values assumed to be the correct inertial manifold dimensions based on other studies \cite{Linot2020,Yang2009,Ding2016}. Here we examine model performance as a function of dimension.

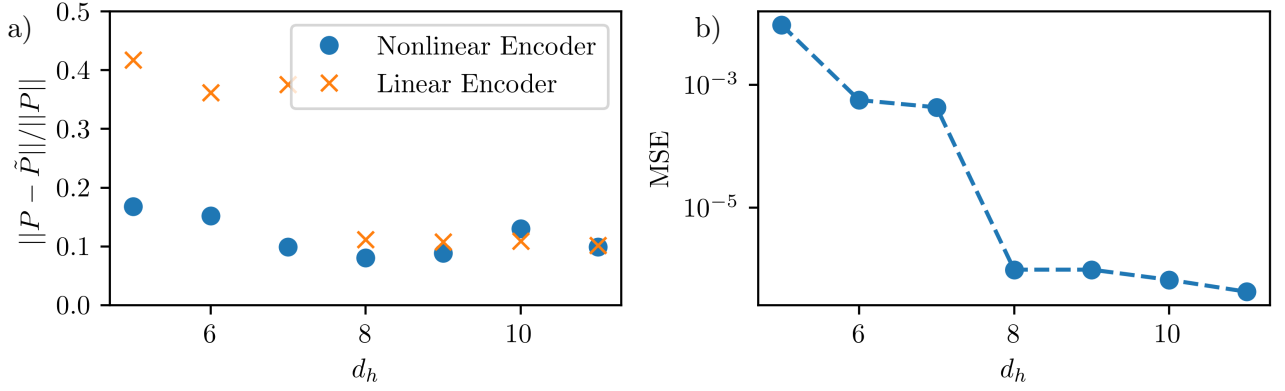
\begin{figure*}
	\centering
	\captionsetup[subfigure]{labelformat=empty}
	\begin{subfigure}[b]{17.2 cm}
		\includegraphics[trim=0 0 0 0,width=\textwidth,clip]{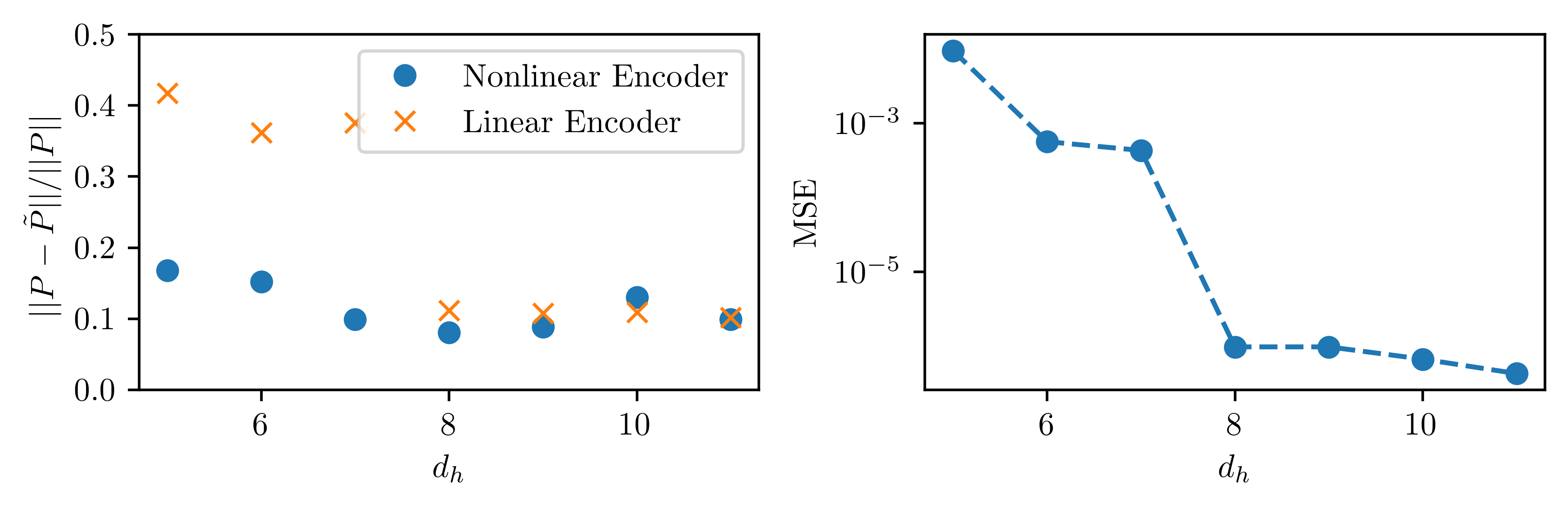}
		\begin{picture}(0,0)
		\put(-240,160){a)}
		\put(20,160){b)}
		\end{picture}
		\caption{}
		\vspace{-10mm}
		\label{fig:Lin_vs_nonlin_PDF}
	\end{subfigure}
	\begin{subfigure}[b]{0\textwidth}\caption{}\vspace{-10mm}\label{fig:Nonlin_Auto}\end{subfigure}
	\captionsetup{justification=raggedright}
	\caption{(a) is the relative error in the joint PDF with the best autoencoder and neural ODE pair at each dimension. (b) is the mean squared error of reconstruction for the nonlinear autoencoder at each dimension. Both are for a domain size of $L=22$.}
	\label{fig:Lin_vs_nonlin}
	\vspace{-5mm}
\end{figure*}

For training we again use $10^5$ time units of data spaced apart 0.25 time units and train 5 autoencoders and 15 neural ODEs. Our first result here is the observation that, for models with a lower dimension than the ``correct" one, reasonable approximations of the joint PDF of $u_x$ and $u_{xx}$ can be obtained, but only if a nonlinear encoder is used for dimension reduction. This point is illustrated in Figure \ref{fig:Lin_vs_nonlin_PDF}, for the case $L=22$, where we see good reconstruction of the joint PDF when using a nonlinear encoder. Accordingly, the results below all use a nonlinear encoder; architectures are reported in Table \ref{Table2}.  However, even though this statistic is reconstructed well at low dimensions, Fig.\ \ref{fig:Nonlin_Auto} shows the mean squared error for the autoencoder, with a nonlinear encoder, still drops significantly at $d_h=8$.

\begin{figure*}
	\centering
	\captionsetup[subfigure]{labelformat=empty}
	\begin{subfigure}[b]{17.2 cm}
		\includegraphics[trim=0 0 0 0,width=\textwidth,clip]{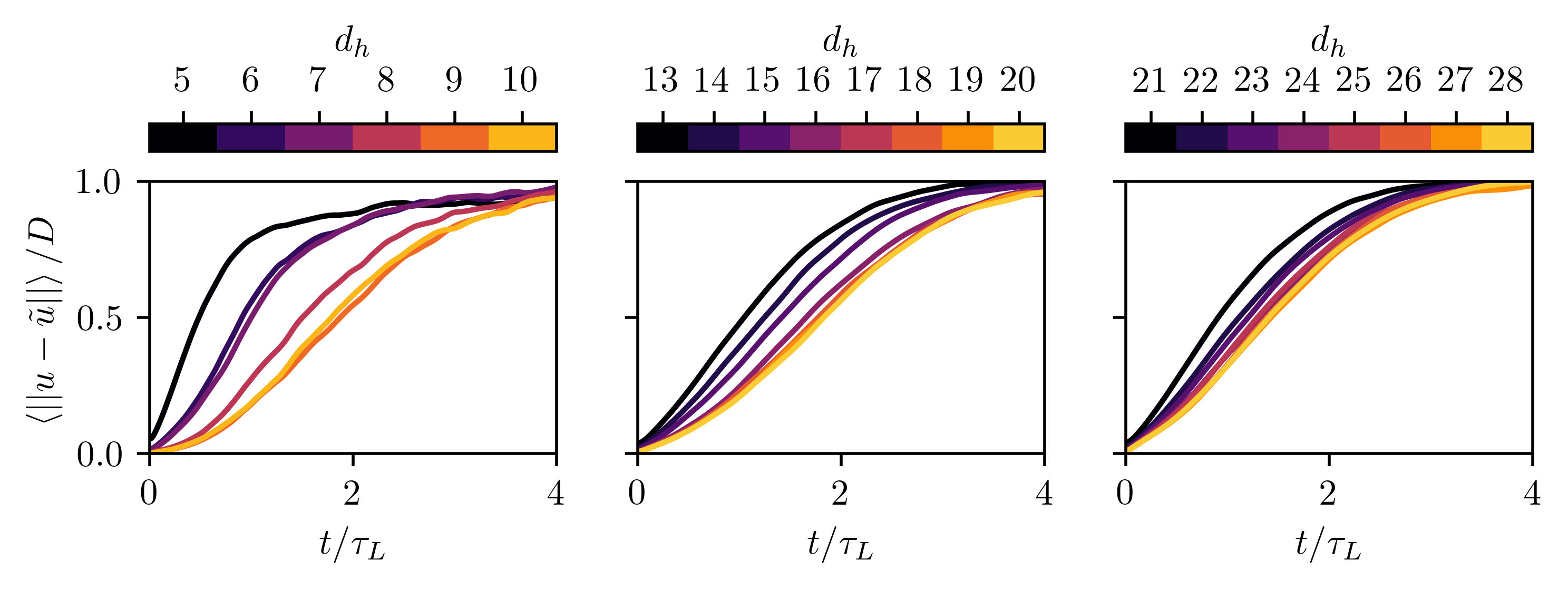}
		\begin{picture}(0,0)
		\put(-235,150){a)}
		\put(-65,150){b)}
		\put(90,150){c)}
		\end{picture}
		\caption{}
		\vspace{-10mm}
		\label{fig:Trajectory_VaryDim}
	\end{subfigure}
	\begin{subfigure}[b]{0\textwidth}\caption{}\vspace{-10mm}\label{fig:Trajectory_VaryDim44}\end{subfigure}
	\begin{subfigure}[b]{0\textwidth}\caption{}\vspace{-10mm}\label{fig:Trajectory_VaryDim66}\end{subfigure}
	\captionsetup{justification=raggedright}
	\caption{Short-time error at various dimensions. Domain sizes are $L=22,44,66$ for (a)-(c).}
	\label{fig:Trajectory_VaryDim_Comb}
	\vspace{-5mm}
\end{figure*}
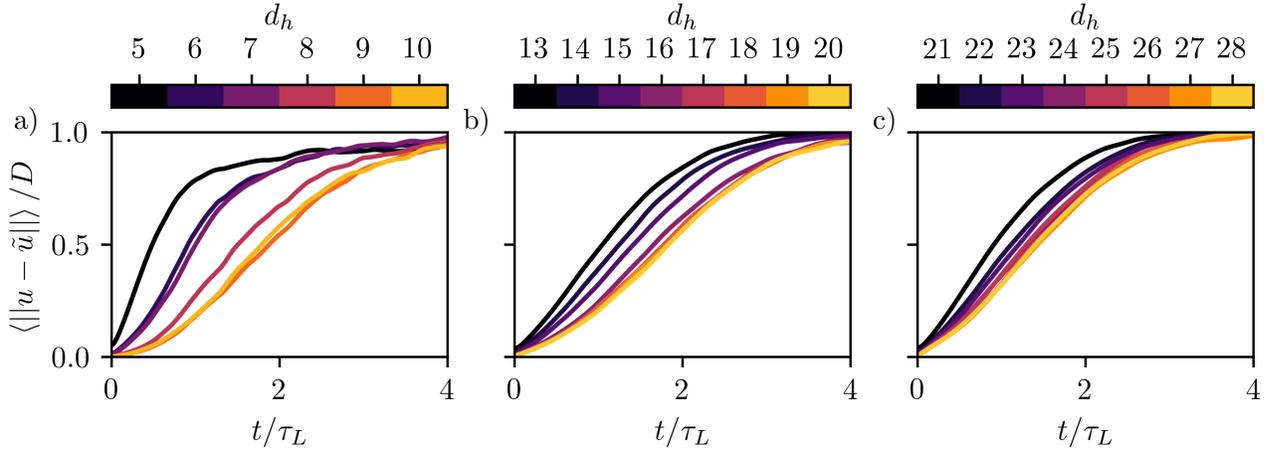   


Figure \ref{fig:Trajectory_VaryDim_Comb} shows the ensemble-averaged short-time tracking error of the models with the best tracking as the dimension varies. 
For all domain sizes the recreation gradually improves and then becomes dimension-independent as dimension increases. This happens because the short-time tracking is directly related to the loss minimized in training the neural ODEs, and $h$ contains the same information as $u$ if $d_h$ is large enough. As mentioned before, for $L=22$, $44$, and $66$, respectively, the ``correct" manifold dimensions are $8$, $18$, and $28$, and Fig.\ \ref{fig:Trajectory_VaryDim_Comb} shows the tracking error becoming dimension-independent near these values. 




%

\begin{figure*}
	\centering
	\captionsetup[subfigure]{labelformat=empty}
	\begin{subfigure}[b]{17.2 cm}
		\includegraphics[trim=0 0 0 0,width=\textwidth,clip]{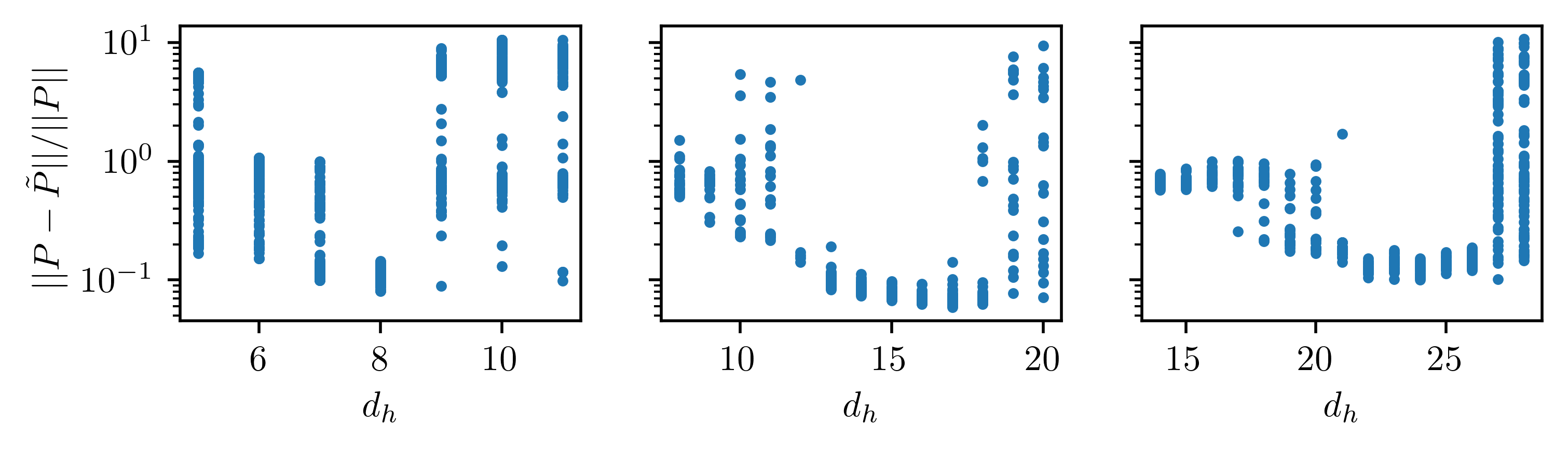}
		\begin{picture}(0,0)
		\put(-230,145){a)}
		\put(-55,145){b)}
		\put(95,145){c)}
		\end{picture}
		\caption{}
		\vspace{-10mm}
		\label{fig:JointPDFError_VaryDim}
	\end{subfigure}
	\begin{subfigure}[b]{0\textwidth}\caption{}\vspace{-10mm}\label{fig:JointPDFError_VaryDim44}\end{subfigure}
	\begin{subfigure}[b]{0\textwidth}\caption{}\vspace{-10mm}\label{fig:JointPDFError_VaryDim66}\end{subfigure}
	\captionsetup{justification=raggedright}
	\caption{Relative error in Joint PDFs for all trained models at different dimensions. Domain sizes are $L=22,44,66$ for (a)-(c).}
	\label{fig:JointPDFError_VaryDim_Comb}
	\vspace{-5mm}
\end{figure*}
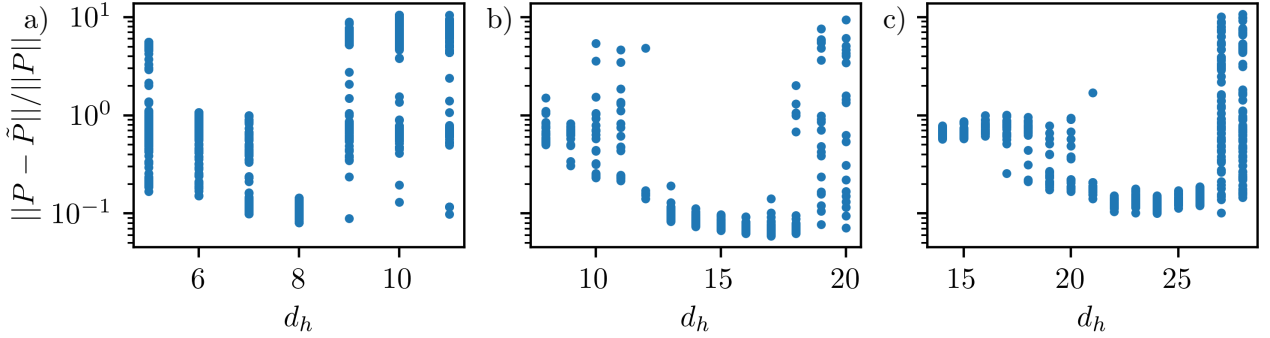

Now we turn to long-time statistical recreation upon varying dimension. Figure \ref{fig:JointPDFError_VaryDim_Comb} shows the relative error in the joint PDFs of $u_x$ and $u_{xx}$ for all the autoencoder and neural ODE pairs trained for each dimension and domain size. Unlike short-time tracking, the relative error of the best models does not monotonically drop with dimension. Also, Fig.\ \ref{fig:JointPDFError_VaryDim_Comb} shows there is a large differences in model performance at a given dimension, despite all models at a given dimension reaching a similar loss during training. These large differences come from trajectories of many of the models leaving the attractor, which can happen because the loss is directly tied to short-time tracking, not long-time statistics. 
In all the cases, when the dimension is low the models do a poor job of reconstructing the joint PDF. When we are near the expected manifold dimension, the average model performance improves and the variance in performance between the models decreases. However, after further increasing the dimension there is an increase in the variance in performance between models, despite little change in the performance of the best models. Like the short-time tracking, the joint PDF of the best models become dimension-independent, but here we see that issues arise when training the neural ODE with unnecessary dimensions.



\section{Conclusion} \label{sec:Conclusion}

Neural ODEs, in conjunction with a low-dimensional approximation of the state in the manifold coordinates, provide a means of accurate data-driven time-evolution near the expected manifold dimension using data that is widely spaced in time. Dimension reduction is a vital step in the process. Not only is training and evolution of Neural ODEs more expensive with more dimensions, but we find that trying to approximate the dynamics with too many dimensions leads to the excitation of high-wavenumber modes that push trajectories off the attractor. With dimension reduction, we find that training with data spaced up to $\sim0.7$ Lyapunov times gives accurate short-time tracking and long-time statistical reconstruction. Finally, we investigate the impact of varying the dimension and find that performance improves with dimension, and then levels off. However, keeping too many dimensions hurts training, resulting in many poor models.


\section*{Data Availability}
The data that support the findings of this study are available from the corresponding author upon reasonable request.

\begin{acknowledgments}
This work was supported by AFOSR  FA9550-18-1-0174 and ONR N00014-18-1-2865 (Vannevar Bush Faculty Fellowship).
\end{acknowledgments}


\begin{thebibliography}{10}

\bibitem{Bengio2004}
Y.~Bengio, J.~F. Paiement, P.~Vincent, O.~Delalleau, N.~{Le Roux}, and
  M.~Ouimet.
\newblock {Out-of-sample extensions for LLE, Isomap, MDS, Eigenmaps, and
  spectral clustering}.
\newblock {\em Advances in Neural Information Processing Systems}, 2004.

\bibitem{Chen2019}
R.~T.~Q. Chen, Y.~Rubanova, J.~Bettencourt, and D.~Duvenaud.
\newblock Neural ordinary differential equations.
\newblock {\em arXiv preprint arXiv:1806.07366}, 2019.

\bibitem{cho-etal-2014-properties}
K.~Cho, B.~van Merri{\"e}nboer, D.~Bahdanau, and Y.~Bengio.
\newblock On the properties of neural machine translation: Encoder{--}decoder
  approaches.
\newblock In {\em Proceedings of {SSST}-8, Eighth Workshop on Syntax, Semantics
  and Structure in Statistical Translation}, pages 103--111, Doha, Qatar, Oct.
  2014. Association for Computational Linguistics.

\bibitem{chollet2015keras}
F.~Chollet et~al.
\newblock Keras.
\newblock \url{https://keras.io}, 2015.

\bibitem{Cunningham2015}
J.~P. Cunningham and Z.~Ghahramani.
\newblock {Linear dimensionality reduction: Survey, insights, and
  generalizations}.
\newblock {\em Journal of Machine Learning Research}, 16:2859--2900, 2015.

\bibitem{ChaosBook}
P.~Cvitanovi{\'c}, R.~Artuso, R.~Mainieri, G.~Tanner, and G.~Vattay.
\newblock {\em Chaos: Classical and Quantum}.
\newblock Niels Bohr Inst., Copenhagen, 2016.

\bibitem{Ding2016}
X.~Ding, H.~Chat{\'{e}}, P.~Cvitanovi{\'{c}}, E.~Siminos, and K.~A. Takeuchi.
\newblock {Estimating the Dimension of an Inertial Manifold from Unstable
  Periodic Orbits}.
\newblock {\em Physical Review Letters}, 117(2):1--5, 2016.

\bibitem{Doering1988}
C.~R. Doering, J.~D. Gibbon, D.~D. Holm, and B.~Nicolaenko.
\newblock {Low-dimensional behaviour in the complex Ginzburg-Landau equation}.
\newblock {\em Nonlinearity}, 1(2):279--309, 1988.

\bibitem{Edson2019}
R.~A. Edson, J.~E. Bunder, T.~W. Mattner, and A.~J. Roberts.
\newblock Lyapunov exponents of the {K}uramoto–{S}ivashinsky {PDE}.
\newblock {\em The ANZIAM Journal}, 61(3):270–285, 2019.

\bibitem{Ferguson2011}
A.~L. Ferguson, A.~Z. Panagiotopoulos, I.~G. Kevrekidis, and P.~G. Debenedetti.
\newblock {Nonlinear dimensionality reduction in molecular simulation: The
  diffusion map approach}.
\newblock {\em Chemical Physics Letters}, 509(1-3):1--11, 2011.

\bibitem{floryan2021charts}
D.~Floryan and M.~D. Graham.
\newblock Charts and atlases for nonlinear data-driven models of dynamics on
  manifolds.
\newblock {\em arXiv preprint arXiv:2108.05928}, 2021.

\bibitem{Foias1988}
C.~Foias, O.~Manley, and R.~Temam.
\newblock Modelling of the interaction of small and large eddies in two
  dimensional turbulent flows.
\newblock {\em ESAIM: Mathematical Modelling and Numerical Analysis -
  Mod\'elisation Math\'ematique et Analyse Num\'erique}, 22(1):93--118, 1988.

\bibitem{Foias1988a}
C.~Foias, B.~Nicolaenko, G.~R. Sell, and R.~Temam.
\newblock {Inertial manifold for the Kuramoto-Sivashinsky equation and an
  estimate of their lowest dimension}.
\newblock {\em J. Math. Pure Appl.}, 67:197--226, 1988.

\bibitem{Fukami2020}
K.~Fukami, T.~Nakamura, and K.~Fukagata.
\newblock Convolutional neural network based hierarchical autoencoder for
  nonlinear mode decomposition of fluid field data.
\newblock {\em Physics of Fluids}, 32(9):095110, 2020.

\bibitem{Gonzalez-Garcia1998}
R.~Gonzalez-Garcia, R.~Rico-Martinez, and I.~G. Kevrekidis.
\newblock Identification of distributed parameter systems: A neural net based
  approach.
\newblock {\em Computers \& Chemical EngineeringComputers \& Chemical
  Engineering}, 22:S965--S968, 1998.

\bibitem{Hinton2003}
G.~Hinton and S.~Roweis.
\newblock Stochastic neighbor embedding.
\newblock {\em Advances in neural information processing systems}, 15:833--840,
  2003.

\bibitem{Hinton2006}
G.~E. Hinton and R.~R. Salakhutdinov.
\newblock Reducing the dimensionality of data with neural networks.
\newblock {\em Science}, 313(5786):504--507, 2006.

\bibitem{Hochreiter1997}
S.~Hochreiter and J.~Schmidhuber.
\newblock {Long Short-Term Memory}.
\newblock {\em Neural Computation}, 9(8):1735--1780, 1997.

\bibitem{IanGoodfellowYoshuaBengio2017}
A.~C. {Ian Goodfellow, Yoshua Bengio}.
\newblock {\em {The Deep Learning Book}}, volume 521.
\newblock 2017.

\bibitem{Ivancevic2007}
V.~G. Ivancevic.
\newblock {\em High-Dimensional Chaotic and Attractor Systems}.
\newblock Springer Publishing Company, Incorporated, 2007.

\bibitem{Jolly2000}
M.~S. Jolly, R.~Rosa, and R.~Temam.
\newblock {Evaluating the dimension of an inertial manifold for the
  Kuramoto-Sivashinsky equation}.
\newblock {\em Advances in Differential Equations}, 5(1-3):31--66, 2000.

\bibitem{Kassam2005}
A.-K. Kassam and L.~N. Trefethen.
\newblock Fourth-order time-stepping for stiff {PDE}s.
\newblock {\em SIAM Journal on Scientific Computing}, 26(4):1214--1233, 2005.

\bibitem{DMDBook}
J.~N. Kutz, S.~L. Brunton, B.~W. Brunton, and J.~L. Proctor.
\newblock {\em Dynamic Mode Decomposition}.
\newblock Society for Industrial and Applied Mathematics, Philadelphia, PA,
  2016.

\bibitem{lee2003}
J.~Lee and J.~Lee.
\newblock {\em Introduction to Smooth Manifolds}.
\newblock Graduate Texts in Mathematics. Springer, 2003.

\bibitem{Linot2020}
A.~J. Linot and M.~D. Graham.
\newblock Deep learning to discover and predict dynamics on an inertial
  manifold.
\newblock {\em Phys. Rev. E}, 101:062209, 2020.

\bibitem{MAULIK2020}
R.~Maulik, A.~Mohan, B.~Lusch, S.~Madireddy, P.~Balaprakash, and D.~Livescu.
\newblock Time-series learning of latent-space dynamics for reduced-order model
  closure.
\newblock {\em Physica D: Nonlinear Phenomena}, 405:132368, 2020.

\bibitem{Milano2002}
M.~Milano and P.~Koumoutsakos.
\newblock {Neural network modeling for near wall turbulent flow}.
\newblock {\em Journal of Computational Physics}, 182(1):1--26, 2002.

\bibitem{Nair2020}
N.~J. Nair and A.~Goza.
\newblock {Leveraging reduced-order models for state estimation using deep
  learning}.
\newblock {\em Journal of Fluid Mechanics}, 897:1--13, 2020.

\bibitem{Omata2019}
N.~Omata and S.~Shirayama.
\newblock A novel method of low-dimensional representation for temporal
  behavior of flow fields using deep autoencoder.
\newblock {\em AIP Advances}, 9(1):015006, 2019.

\bibitem{Page2020}
J.~Page, M.~P. Brenner, and R.~R. Kerswell.
\newblock Revealing the state space of turbulence using machine learning.
\newblock {\em Phys. Rev. Fluids}, 6:034402, 2021.

\bibitem{Paszke2019}
A.~Paszke, S.~Gross, F.~Massa, A.~Lerer, J.~Bradbury, G.~Chanan, T.~Killeen,
  Z.~Lin, N.~Gimelshein, L.~Antiga, A.~Desmaison, A.~Kopf, E.~Yang, Z.~DeVito,
  M.~Raison, A.~Tejani, S.~Chilamkurthy, B.~Steiner, L.~Fang, J.~Bai, and
  S.~Chintala.
\newblock Pytorch: An imperative style, high-performance deep learning library.
\newblock In {\em Advances in Neural Information Processing Systems 32}, pages
  8024--8035. Curran Associates, Inc., 2019.

\bibitem{Pathak2018a}
J.~Pathak, B.~Hunt, M.~Girvan, Z.~Lu, and E.~Ott.
\newblock {Model-Free Prediction of Large Spatiotemporally Chaotic Systems from
  Data: A Reservoir Computing Approach}.
\newblock {\em Physical Review Letters}, 120(2):24102, 2018.

\bibitem{portwood2019turbulence}
G.~D. Portwood, P.~P. Mitra, M.~D. Ribeiro, T.~M. Nguyen, B.~T. Nadiga, J.~A.
  Saenz, M.~Chertkov, A.~Garg, A.~Anandkumar, A.~Dengel, R.~Baraniuk, and D.~P.
  Schmidt.
\newblock Turbulence forecasting via neural {ODE}.
\newblock {\em arXiv preprint arXiv:1911.05180}, 2019.

\bibitem{Raissi2018}
M.~Raissi, P.~Perdikaris, and G.~E. Karniadakis.
\newblock {Multistep Neural Networks for Data-driven Discovery of Nonlinear
  Dynamical Systems}.
\newblock {\em arXiv preprint arXiv:1801.01236}, pages 1--19, 2018.

\bibitem{rojas2021reducedorder}
C.~J.~G. Rojas, A.~Dengel, and M.~D. Ribeiro.
\newblock Reduced-order model for fluid flows via neural ordinary differential
  equations.
\newblock {\em arXiv preprint arXiv:2102.02248}, 2021.

\bibitem{Roweis2323}
S.~T. Roweis and L.~K. Saul.
\newblock Nonlinear dimensionality reduction by locally linear embedding.
\newblock {\em Science}, 290(5500):2323--2326, 2000.

\bibitem{Sauer1991}
T.~Sauer, J.~A. Yorke, and M.~Casdagli.
\newblock Embedology.
\newblock {\em Journal of Statistical Physics}, 65(3):579--616, 1991.

\bibitem{Takens}
F.~Takens.
\newblock Detecting strange attractors in turbulence.
\newblock In D.~Rand and L.-S. Young, editors, {\em Dynamical Systems and
  Turbulence, Warwick 1980}, pages 366--381, Berlin, Heidelberg, 1981. Springer
  Berlin Heidelberg.

\bibitem{Temam1989}
R.~Temam.
\newblock {Do inertial manifolds apply to turbulence?}
\newblock {\em Physica D: Nonlinear Phenomena}, 37(1-3):146--152, 1989.

\bibitem{Temam1990}
R.~Temam.
\newblock {Inertial manifolds}.
\newblock {\em The Mathematical Intelligencer}, 12(4):68--74, 1990.

\bibitem{Temam1994}
R.~Temam and X.~Wang.
\newblock {Estimates on the lowest dimension of inertial manifolds for the
  Kuramoto-Sivasbinsky equation in the general case}.
\newblock {\em Differential and Integral Equations}, 7(3-4):1095--1108, 1994.

\bibitem{VanDerMaaten2009}
L.~J.~P. {Van Der Maaten}, E.~O. Postma, and H.~J. {Van Den Herik}.
\newblock {Dimensionality Reduction: A Comparative Review}.
\newblock {\em Journal of Machine Learning Research}, 10:1--41, 2009.

\bibitem{Vlachas2019}
P.~Vlachas, J.~Pathak, B.~Hunt, T.~Sapsis, M.~Girvan, E.~Ott, and
  P.~Koumoutsakos.
\newblock Backpropagation algorithms and reservoir computing in recurrent
  neural networks for the forecasting of complex spatiotemporal dynamics.
\newblock {\em Neural Networks}, 126:191--217, 2020.

\bibitem{Vlachas2018}
P.~R. Vlachas, W.~Byeon, Z.~Y. Wan, T.~P. Sapsis, and P.~Koumoutsakos.
\newblock Data-driven forecasting of high-dimensional chaotic systems with long
  short-term memory networks.
\newblock {\em Proceedings of the Royal Society A: Mathematical, Physical and
  Engineering Sciences}, 474(2213):20170844, 2018-05.

\bibitem{Whitney1944}
H.~Whitney.
\newblock The self-intersections of a smooth n-manifold in 2n-space.
\newblock {\em Annals of Mathematics}, 45(2):220--246, 1944.

\bibitem{Yang2009}
H.~L. Yang, K.~A. Takeuchi, F.~Ginelli, H.~Chat{\'{e}}, and G.~Radons.
\newblock {Hyperbolicity and the effective dimension of spatially extended
  dissipative systems}.
\newblock {\em Physical Review Letters}, 102(7):1--4, 2009.

\bibitem{Zelik2014}
S.~Zelik.
\newblock {Inertial manifolds and finite-dimensional reduction for dissipative
  PDEs}.
\newblock {\em Proceedings of the Royal Society of Edinburgh Section A:
  Mathematics}, 144(6):1245--1327, 2013.

\end{thebibliography}
\end{document}